\pdfoutput=1

\documentclass[11pt]{article}
\PassOptionsToPackage{table}{xcolor}
\usepackage[final]{acl}
\usepackage{CJKutf8}
\usepackage[utf8]{inputenc}
\usepackage{times}
\usepackage{latexsym}
\usepackage{amsmath}
\usepackage{subcaption}
\usepackage{multirow}
\usepackage{booktabs}
\usepackage[table, dvipsnames]{xcolor}	
\usepackage{tcolorbox}
\usepackage{arydshln}
\usepackage{amssymb}
\usepackage{float}
\usepackage{tabularx}
\usepackage{makecell}
\usepackage{colortbl}
\usepackage{array}
\usepackage{algpseudocode}

\usepackage[ruled,linesnumbered,vlined]{algorithm2e}
\usepackage{xcolor}
\usepackage{xspace}

\definecolor{searchblue}{RGB}{0, 176, 240}
\definecolor{answerred}{RGB}{255, 0, 102}
\definecolor{infogold}{RGB}{255, 153, 0}

\usepackage[T1]{fontenc}

\usepackage[utf8]{inputenc}

\usepackage{microtype}

\usepackage{inconsolata}

\usepackage{graphicx}

\usepackage[normalem]{ulem}
\usepackage{fontawesome5} 
\usepackage{enumitem}

\newcommand{\ours}{\textproc{LcRL}}
\title{Language-Coupled Reinforcement Learning for Multilingual Retrieval-Augmented Generation}

\author{
 \textbf{Rui Qi\textsuperscript{1,2}},
 \textbf{Fengran Mo\textsuperscript{3}},
 \textbf{Yufeng Chen\textsuperscript{1,2}\footnotemark[2]},
 \textbf{Xue Zhang\textsuperscript{1,2}},
 \textbf{Shuo Wang\textsuperscript{4}},
 \textbf{Hongliang Li\textsuperscript{1,2}}\\
 \textbf{Jinan Xu\textsuperscript{1,2}},
 \textbf{Jian-Yun Nie\textsuperscript{3}},
\textbf{Kaiyu Huang\textsuperscript{1,2}\footnotemark[2]}
\\
 \textsuperscript{1}Key Laboratory of Big Data \& Artificial Intelligence in Transportation \\
 （Beijing Jiaotong University), Ministry of Education \\
\textsuperscript{2}School of Computer Science and Technology, Beijing Jiaotong University \qquad \\
 \textsuperscript{3}University of Montreal \qquad
 \textsuperscript{4}Tsinghua University \qquad
 \texttt{\{cherry,kyhuang\}@bjtu.edu.cn}
}

\begin{document}
\begin{CJK*}{UTF8}{gkai}
\maketitle
\renewcommand{\thefootnote}{\fnsymbol{footnote}}
\footnotetext[2]{Corresponding author.}
\begin{abstract}
Multilingual retrieval-augmented generation~(MRAG) requires models to effectively acquire and integrate beneficial external knowledge from multilingual collections.
However, most existing studies employ a unitive process where queries of equivalent semantics across different languages are processed through a single-turn retrieval and subsequent optimization.
Such a ``one-size-fits-all'' strategy is often suboptimal in multilingual settings, as the models occur to knowledge bias and conflict during the interaction with the search engine.
To alleviate the issues, we propose \ours, a multilingual search-augmented reinforcement learning framework that integrates a language-coupled Group Relative Policy Optimization into the policy and reward models.
We adopt the language-coupled group sampling in the rollout module to reduce knowledge bias, and regularize an auxiliary anti-consistency penalty in the reward models to mitigate the knowledge conflict.
Experimental results demonstrate that \ours~not only achieves competitive performance but is also appropriate for various practical scenarios such as constrained training data and retrieval over collections encompassing a large number of languages. Our code is available at \url{https://github.com/Cherry-qwq/LcRL-Open}.
\end{abstract}


\section{Introduction}
Retrieval-augmented generation~(RAG) has emerged as a powerful solution to incorporate external knowledge into Large Language Models~(LLMs) to enhance the performance via retrieving supportive evidence 
for knowledge-intensive tasks, which alleviate the hallucinations and lack of knowledge in training corpora~\cite{ni-etal-2024-llms, ding2025rowenadaptiveretrievalaugmentedgeneration,zhang-etal-2025-cm}.
In multilingual settings, the training data of LLMs is notoriously imbalanced and culturally biased, leading to significant knowledge disparities across languages~\cite{huang2024survey,ranaldi-etal-2025-improving-multilingual,zhang-etal-2025-multilingual,man-etal-2025-dmdteval}. Consequently, RAG remains essential, as it enables LLMs to leverage diverse multilingual collections to bridge these information gaps and enhance the models' capabilities in multiple languages.

\begin{figure}[!t]
    \centering
    \includegraphics[width=\linewidth]{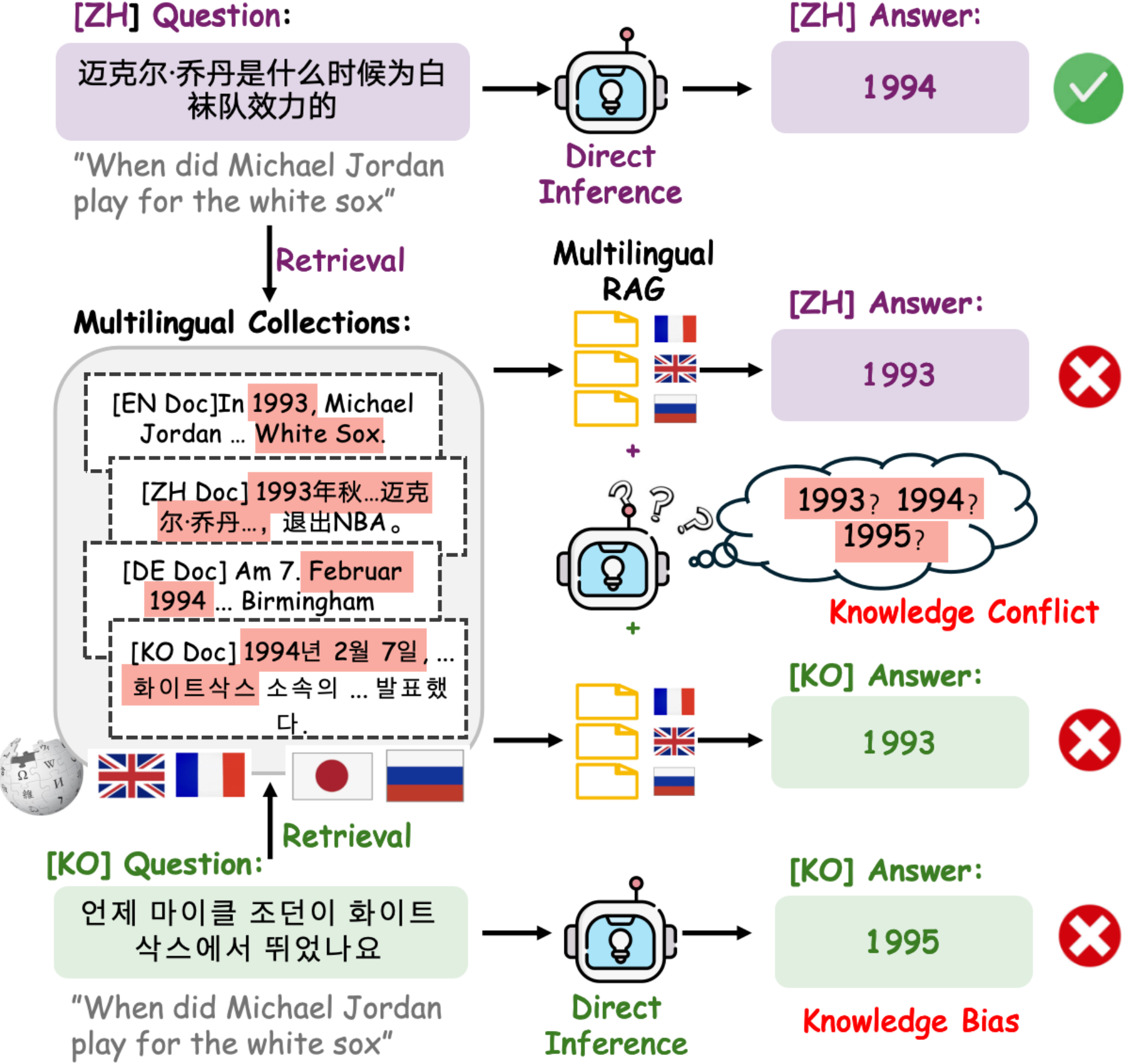}
    \caption{Illustration of knowledge bias and knowledge conflict challenges in the MRAG scenario.}
    \label{fig:gap}
\end{figure}

Existing approaches in multilingual retrieval-augmented generation~(MRAG) typically identify relevant passages from the multilingual collections for a given query as a single-turn retrieval process~\cite{ranaldi2025multilingualretrievalaugmentedgenerationknowledgeintensive, wu2024languagesequalinsightsmultilingual, kumar-etal-2025-bridging, ranaldi-etal-2025-improving-multilingual, park-lee-2025-investigating}. 
Although showing effectiveness,
the methods encounter two challenges~\cite{reusens2023investigating,chirkova2024retrieval}: \textit{knowledge bias} and \textit{knowledge conflict}. 
As shown in Figure~\ref{fig:gap}, LLMs might generate divergent responses to the identical queries expressed in different languages, due to inherent disparities in knowledge and capabilities specific to each language, a phenomenon referred to as knowledge bias. 
On the other hand, when the retrieval collections contain a wide variety of languages, differences in linguistic expression lead to retrieved documents that are semantically related but factually inconsistent, resulting in knowledge conflicts that hinder the LLM from producing correct responses.
Based on these observations, a research question arises: Can LLMs adaptively determine the need for retrieving external knowledge and effectively harmonize contradictory knowledge obtained from different languages, particularly when dealing with a diverse and extensive set of multilingual collections?


To address these issues, the development of reinforcement learning~(RL) based post-training provides the possibility to enable the LLMs achieve autonomous query generation and dynamic utilization of retrieved information through optimizing its rollout process~\cite{jin2025searchr1trainingllmsreason,wu2025structurer1dynamicallyleveragingstructural,lupart2025chatr1,mo2026agentic}.
In these efforts, LLMs can interact with search engines in an interleaved manner and support multi-turn reasoning.
Such a paradigm provides the potential idea to dynamically adjust the retrieval and response generation strategies based on the complexity of queries in different languages among MRAG scenarios.

In this study, considering the complementary effect among languages, we propose a language-coupled reinforcement learning~(\ours) method that couples multilingual decision-making and experience reward into the search-augmented RL framework, aiming to mitigate the issues of knowledge bias and knowledge conflict in MRAG. 
In particular, \ours~adopts Group Relative Policy Optimization (GRPO)~\cite{deepseekai} with a language-coupled rollout mechanism, which mitigates the knowledge bias of multilingual queries through group scores, and breaks the language-independent pattern to address knowledge conflicts arising from factual inconsistencies in single-language queries over multilingual collections.
For semantically equivalent queries, the language-coupled sampling generation within the same group focuses on optimizing the model's ability to interact with multilingual collections and perform reasoning in a multi-turn and multilingual paradigm.
As a result, queries in different languages can adapt to their respective external retrieval requirements and implicitly learn the transferability from the behaviors of the other languages.

In terms of RL-based model training, we refine the outcome-based reward function by incorporating an $n$-gram-based recall score, which improves adaptability for multilingual generations of varying lengths. We observe that while the language-coupled GRPO framework exhibits robust cross-lingual transfer capabilities, it encounters significant training instability as the number of supported languages increases.
To alleviate this issue, we introduce an auxiliary anti-consistency penalty regularization, which stabilizes the training process by assessing the correctness of the language-coupled response. 
Experimental results show that our method effectively accommodates increasing language diversity in the retrieval collections and exhibits strong performance and robustness under limited training data conditions.
To sum up, our contributions are as follows:
\begin{itemize}
    \item We propose \ours, a multilingual search-augmented GRPO framework that is integrated with a language-coupled rollout module and reward modeling, achieving competitive performance on MRAG compared with several baselines.
    \item Our proposed multi-turn retrieval strategy with the auxiliary anti-consistency penalty regularization provides adaptive control of search needs in multilingual settings and exhibits enhanced performance with increasing language counts.
    \item Experiments show that \ours~performs strong generalization, which is minimally influenced by the scale and language diversity of training data, successfully transfers to unseen languages, and adapts various LLMs.
    
\end{itemize}

\begin{figure*}[!t]
    \centering
    \includegraphics[width=1\linewidth]{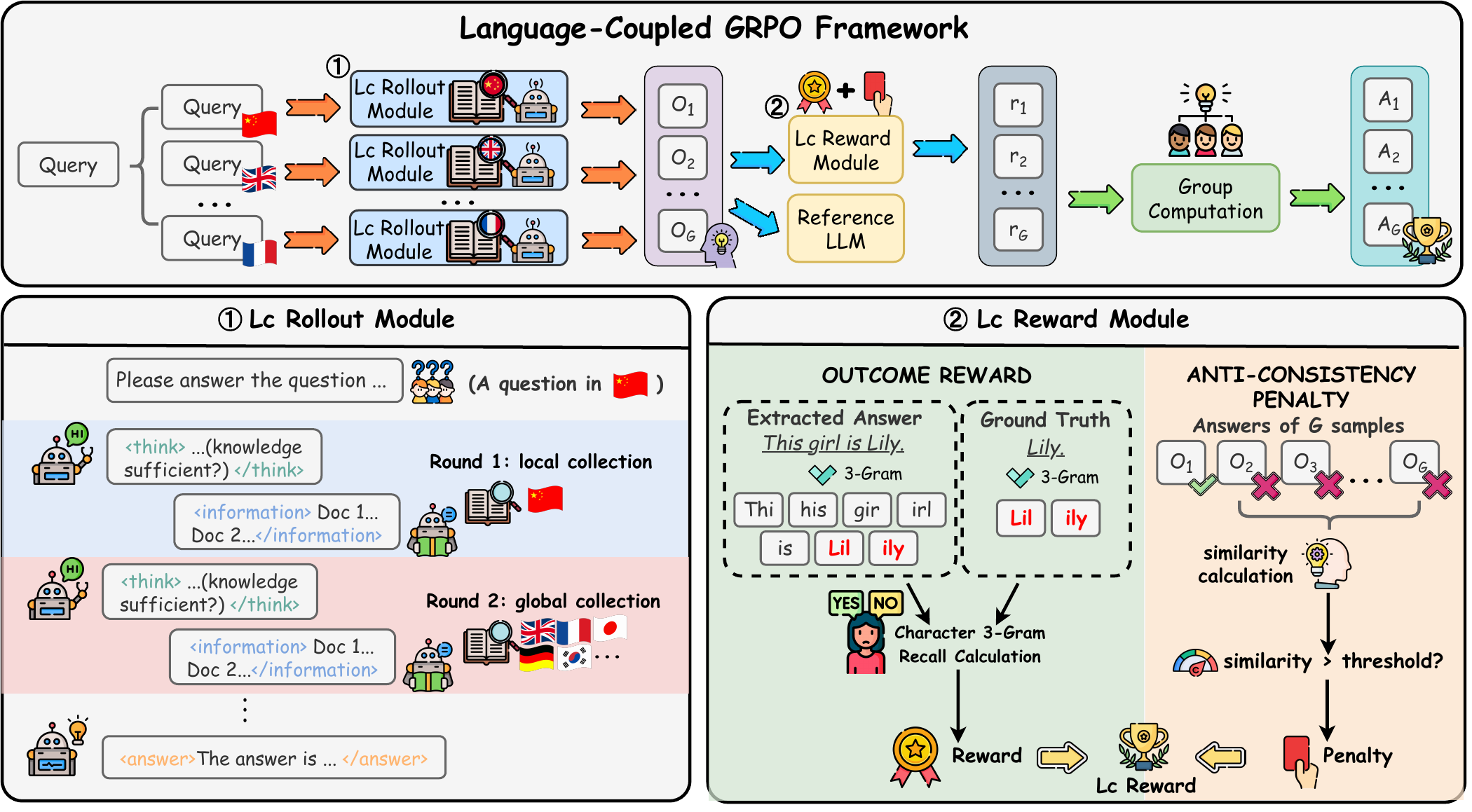}
    \caption{Illustration of our \ours by integrating two modules (rollout and reward) within the GRPO for language-coupled (Lc) purposes. 
    The Lc rollout is designed with sampling in multiple languages for multilingual scenarios. The Lc reward is incorporated with an anti-consistency mechanism to mitigate training collapse.
    }
    \label{fig:placeholder}
    \vspace{-2ex}
\end{figure*}

\section{Related Work}
\paragraph{Multilingual RAG.} Early research in MRAG focused mainly on the empirical optimization of individual components within the pipeline, such as the introduction of translation strategies ~\cite{chirkova2024retrieval, park-lee-2025-investigating, wu2024languagesequalinsightsmultilingual, ranaldi2025multilingualretrievalaugmentedgenerationknowledgeintensive,zhang-etal-2025-less}.
The other studies attempt to maximize the efficacy of LLMs across languages through dynamic adaptation ~\cite{kumar-etal-2025-bridging}, including incorporating multi-step dialectic reasoning to alleviate the knowledge conflict during inference ~\cite{ranaldi-etal-2025-improving-multilingual}.
However, existing approaches predominantly operate within a fixed pipeline framework, improving performance through point-wise modifications of isolated modules ~\cite{chirkova2024retrieval, 10594504}.
In contrast to these methods, our approach jointly optimizes retrieval and generation, which dynamically governs the retrieval strategy based on the inherent knowledge bias required for generation across different languages, while effectively mitigating knowledge conflicts in MRAG.
\paragraph{Reinforcement Learning for RAG.}

The integration of external tools has evolved from static RAG to dynamic, agentic workflows. 
Such methods, represented by Search-R1~\cite{jin2025searchr1trainingllmsreason}, allow LLMs to autonomously generate multiple search queries, interact with search engines in a multi-step manner, and perform logical inference purely driven by reward signals ~\cite{kumar2024training,chen2025researchlearningreasonsearch,deepseekai,man-etal-2024-icl, zhang2025ratt,zhang2026thinknativelyunlockingmultilingual}
Building upon this paradigm, recent studies have significantly expanded the capabilities of search-augmented agents~\cite{zhang2024blind,luo2025maragr1singleretrieverreinforcementlearned,song2025r1searcherincentivizingdynamicknowledge,lin2025resistingcontextualinterferencerag,mo2026opendecoder}, including efficiency~\cite{xu2025reconreasoningcondensationefficient,lin2025resistingcontextualinterferencerag} and structural comprehension~\cite{wu2025structurer1dynamicallyleveragingstructural,wu-etal-2025-graph}. Different from monolingual scenarios, multilingual settings exhibit knowledge conflicts and biases arising from multilingual capabilities.
In such situations, current RL-based policies fail to reconcile the contradictory facts yielded by semantically equivalent queries across languages. To address this issue, we propose a language-coupled RL-based framework to solve knowledge bias and knowledge conflict in the rollout and reward phases, respectively.

\section{Method}

In monolingual scenarios, knowledge bias typically refers to systematic preferences induced by data distribution and model priors, while knowledge conflict arises from mutually inconsistent evidence within the same language corpus. However, the situation is different in RAG. In particular, \textbf{knowledge bias} stems from inherent disparities in training data across languages, leading to divergent reasoning capabilities even for semantically equivalent queries.
\textbf{Knowledge conflict} occurs when top-ranked passages from a multilingual collection present factually inconsistent information, misleading the LLM due to information overload and cross-lingual contradictions.
To address specific issues of MRAG, \ours~optimizes reasoning trajectories by coupling multilingual decision-making and experience rewards within a unified RL framework.

\subsection{Response Generation with Multi-Turn and Multilingual Retrieval}
\label{sec:multi-turn}
The rollout process for LLM response generation using an interleaved, multi-turn retrieval mechanism designed for the multilingual scenarios is described in Algorithm~\ref{alg:multilingual_rollout}.
We define the response generation process is as:$$y \sim \pi_{\theta}(\cdot | q_L; \mathcal{R}_{multi}) = \pi_{\theta}(\cdot | q_L) \otimes \{\mathcal{R}_l\}_{l \in \mathcal{L}}$$where $q_L$ is the input query with the source language $L$, and $\mathcal{R}_{multi} = \{\mathcal{R}_l\}_{l \in \mathcal{L}}$ represents the set of all language-specific retrievers for each language $l$ in the multilingual collection $\mathcal{L}$.
The model executes a <think> phase to 
activate the parametric memory. If internal knowledge is insufficient, the model executes a hierarchical retrieval strategy by searching useful information from different language collections.

\begin{algorithm}[!t]
\SetAlFnt{\small}
\SetAlCapFnt{\small}
\SetAlCapNameFnt{\small}
\tiny
\caption{Multilingual Multi-Turn Retrieval-Augmented Generation}
\label{alg:multilingual_rollout}
\DontPrintSemicolon

\SetKwInOut{Require}{Require}
\SetKwInOut{Ensure}{Ensure}
\Require{Input query $q_L$, source language $L$, policy model $\pi_\theta$, multilingual retriever set $\mathcal{R}_{multi} = \{\mathcal{R}_l\}_{l \in \mathcal{L}}$, maximum action budget $B$.}
\Ensure{Final multilingual-informed response $y$.}

Initialize rollout sequence $y \leftarrow \emptyset$\;
Initialize action count $b \leftarrow 0$\;

\While{$b < B$}{
    Initialize current action rollout sequence $y_b \leftarrow \emptyset$\;
    \While{True}{
        Generate response token $y_t \sim \pi_\theta(\cdot \mid q_L, y + y_b)$\;
        Append $y_t$ to rollout sequence $y_b \leftarrow y_b + y_t$\;
        \If{$y_t \in [ \text{\textcolor{searchblue}{</search>}}, \text{\textcolor{answerred}{</answer>}}, \text{<eos>} ]$}{
            \textbf{break}\;
        }
    }
    $y \leftarrow y + y_b$\;
    
    \If{\textcolor{searchblue}{<search> </search>} detected in $y_b$}{
        Extract search query $q \leftarrow \text{Parse}(y_b, \textcolor{searchblue}{<search>}, \textcolor{searchblue}{</search>})$\;
        Increment action count $b \leftarrow b + 1$\;
        
        \tcp{Hierarchical Retrieval Strategy}
        \If{$b = 1$}{
            \tcp{Turn 1: Native Language Retrieval}
            Retrieve results $d = \mathcal{R}_L(q)$\;
        }
        \ElseIf{$b = 2$}{
            \tcp{Turn 2: Global Knowledge Expansion}
            Retrieve results $d = \bigcup_{l \in \mathcal{L} \setminus \{L\}} \mathcal{R}_l(q)$\;
        }
        \Else{
            \tcp{Turn $\ge 3$: High-Resource Anchoring}
            Retrieve results $d = \mathcal{R}_{en}(q)$\;
        }
        
        Insert $d$ into rollout $y \leftarrow y + \text{\textcolor{infogold}{<information>}} d \text{\textcolor{infogold}{</information>}}$\;
    }
    \ElseIf{\textcolor{answerred}{<answer> </answer>} detected in $y_b$}{
        \Return{final generated response $y$}\;
    }
    \Else{
        Ask for rethink $y \leftarrow y + \text{``My action is not correct. Let me rethink.''}$\;
    }
}
\Return{final generated response $y$}\;
\end{algorithm}
\textbf{\texttt{Turn 1: Native Language Retrieval.}} The model first invokes the retriever $\mathcal{R}_L$ corresponding specifically to the query's native language $L$, which can capture local context and avoid knowledge conflict by leveraging the culturally and contextually relevant evidence.

\textbf{\texttt{Turn 2: Global Knowledge Expansion.}} After processing the native results within a second <think> block, if the model identifies remaining gaps caused by knowledge bias, it triggers a second retrieval turn. The model simultaneously queries all other language collections in the set $\mathcal{L} \setminus \{L\}$, which is formalized as $\mathcal{D}_{global} = \bigcup_{l \in \mathcal{L} \setminus \{L\}} \mathcal{R}_{l}(q_L)$.

\textbf{\texttt{Turn $\ge$ 3: High-Resource Anchoring.}}
For any additional turns, the model is navigated by default to a high-resource retriever~(e.g., English $\mathcal{R}_{en}$) to serve as a factual anchor until an <answer> is produced or the maximum action budget $B$ is consumed. 
This is determined by the broader information coverage and extensive corpus scale in terms of resource~\cite{park-lee-2025-investigating}.

Our multi-turn strategy empowers the model to assess its knowledge adequacy at each step autonomously. 
By prioritizing the native language in the initial turn, the model seeks to minimize information redundancy and circumvent potential conflicts.
The model makes an adaptive decision to expand the search to multilingual, ensuring that cross-lingual complementarity is utilized to address core information deficits, while maintaining the center of native language knowledge to reduce the interference caused by redundant global data. 
The detailed prompting template used for training and interaction is provided in Appendix~\ref{app:template}.

\definecolor{tagblue}{RGB}{0, 102, 204}
\definecolor{tagpurple}{RGB}{128, 0, 128}
\definecolor{taggreen}{RGB}{0, 128, 0}
\definecolor{tagorange}{RGB}{220, 150, 60}
\definecolor{bglight}{RGB}{245, 245, 250}

\subsection{Language-Coupled GRPO with the Search Engine Calling}
As shown in the left part of Figure~\ref{fig:placeholder}, to accommodate multilingual scenarios, we extend the GRPO framework by implementing a language-coupled rollout mechanism. This approach modifies the sampling process to explicitly mitigate linguistic bias and multilingual optimization conflicts.

\paragraph{Multilingual Group Sampling.} 
For a specific question $q$, we construct a set of semantically equivalent queries across multiple languages: $\mathcal{Q} = \{q_{1}, q_{2}, \dots, q_{i}, \dots, q_{n}\}$, where $q_{i}$ represents the corresponding language query.
The multilingual queries originate from the parallel data expressed in other languages, which are usually unavailable in the datasets but can be generated using translation thinking strategies~\cite{qi2025sot}.
After the interaction with the multilingual and multi-turn retrieval, for a group of size $G$, we sample responses $o_i$ such that each $o_i$ corresponds to a different language query from $\mathcal{Q}$ as:
$$o_i \sim \pi_{\theta}(\cdot | q_i; \mathcal{R}), \quad q_i \in \mathcal{Q}$$

\paragraph{Objective Function. }
Distinct from standard GRPO, which optimizes based on a single-query group, our objective operates over the coupled query set $\mathcal{Q}$. 
To mitigate the bias in LLMs that leads to divergent responses to equivalent queries due to inherent language-specific differences, the advantage term $\hat{A}_{i,t}^{\text{coupled}}$ is computed over $\mathcal{Q}$, leveraging a shared baseline derived from the entire multilingual group.
By normalizing the reward of each language-specific response $o_i$ against the collective performance of the group, the policy is encouraged to bind embeddings from different languages to a unified, shared, and high-quality reasoning path, thereby optimizing trajectories within the same group.



Formally, we optimize the policy $\pi_{\theta}$ by maximizing the following language-coupled objective:

$$\small
\begin{aligned}
&\mathcal{J}_{\text{LcRL}}(\theta) = \mathbb{E}_{\{q_i\}_{i=1}^G \sim \mathcal{D}_{\text{couple}}, \{o_i\}_{i=1}^G \sim \pi_{\text{old}(\cdot|\{q_i\}; \mathcal{R})}} \\
&\Bigg[ \frac{1}{G} \sum_{i=1}^G \frac{1}{\sum_{t=1}^{|o_i|} I(o_{i,t})} \sum_{t=1, I(o_{i,t})=1}^{|o_i|} \\
&\min \Bigg( \frac{\pi_{\theta}(o_{i,t}|q_i, o_{i,<t}; \mathcal{R})}{\pi_{\text{old}}(o_{i,t}|q_i, o_{i,<t}; \mathcal{R})} \hat{A}_{i,t}^{\text{coupled}}, 
\text{clip} \\
&\left( \frac{\pi_{\theta}(o_{i,t}|q_i, o_{i,<t}; \mathcal{R})}{\pi_{\text{old}}(o_{i,t}|q_i, o_{i,<t}; \mathcal{R})}, 1-\epsilon, 1+\epsilon \right) \hat{A}_{i,t}^{\text{coupled}} \Bigg) \\
&- \beta \frac{1}{G} \sum_{i=1}^G \mathbb{D}_{\text{KL}} [\pi_{\theta}(\cdot|q_i) || \pi_{\text{ref}}(\cdot|q_i)] \Bigg]
\end{aligned}$$
where $\epsilon$ and $\beta$ are hyperparameters. Additionally, the indicator function $I(o_{i,t})$ masks retrieved tokens, ensuring that gradients are backpropagated solely through the model's generated reasoning and search actions.

\subsection{Language-Coupled Reward Modeling}
To guide the optimization process in search-augmented RL, previous methods~(e.g., Search-R1) adopt the rule-based outcome reward system using exact matching (EM). However, this discrete approach is suboptimal for multilingual tasks where lexical diversity is prevalent. More critically, recent investigations~\cite {deng2025grpocollapsesearchr1lazy} reveal that GRPO-based tool-integrated RL is uniquely vulnerable to training collapse triggered by Lazy Likelihood Displacement (LLD), which is caused by high representational overlap between incorrect and correct trajectories. Thus, we propose the Language-Coupled Reward Modeling system.

\paragraph{Multilingual Outcome Reward.} As word-level matching fails to capture such lexical similarities, we replace the binary EM reward with Character 3-gram Recall ($r_{\text{ans}}$) ~\cite{chirkova2024retrieval} to assess answer correctness across multilingual generations:$$r_{\text{ans}}(i) = \text{c3Recall}(\hat{a}_i, a_{\text{gold}})$$where $\hat{a}_i$ is the extracted answer and $a_{\text{gold}}$ is the ground truth. Unlike EM, $r_{\text{ans}}$ provides a dense reward signal, which slows down the entry into ``high-gradient danger zones'' by stabilizing importance ratios, preventing the advantages from being amplified by extreme reward disparities.

\paragraph{Auxiliary Anti-Consistency Penalty.}
In RL, training instability often arises due to the high variance of the reward signal and the non-stationarity of the environment. To ensure training stability and resolve knowledge conflict, we introduce the Anti-Align Penalty, which acts as a repulsive barrier 
to break the loop of error mode collapse.
We first define the set of ``bad samples'' $B_q$ for a query $q$ as those failing a correctness threshold $\tau_{\text{bad}}$:$$B_q = \{i \in G_q \mid r_{\text{ans}}(i) < \tau_{\text{bad}}\}$$
For each sample $i \in B_q$, we compute its maximum similarity to other incorrect responses in the group: $$m_i = \max_{j \in B_q, j \neq i} \text{Sim}_{c3}(\hat{a}_i, \hat{a}_j), $$
where Sim$_{c3}$ represents the similarity calculated by character 3-gram recall. The base penalty $p_i$ is defined with a margin $\gamma = 0.5$ as $p_i = \max(0, m_i - \gamma)$. To ensure stability, we introduce a sample-level weight $w_q = \min(1, \frac{|B_q|}{5})$ and formulate the anti-alignment reward:$$r_{\text{anti\_align}}(i) = 
\begin{cases} 
-p_i \cdot w_q, & i \in B_q \text{ and } |B_q| > 1 \\
0, & \text{otherwise} 
\end{cases}$$

By penalizing clusters of similar incorrect answers, the Anti-Consistency Penalty incentivizes the model to critically compare inconsistent evidence across languages, and alleviate the negative gradient issue caused by incorrect mode collapse and representational overlap with correct paths. 

\paragraph{Reward Fusion and Normalization.}
To prevent the penalty from dominating the training, we apply a clipping operation as:
$$\tilde{r}_{\text{anti\_align}}(i) = \text{clip}(r_{\text{anti\_align}}(i), -0.5, 0)$$ 
Finally, the final reward is defined as:
$$r_{\text{total}}(i) = \max(0, r_{\text{ans}}(i) + \lambda \cdot \tilde{r}_{\text{anti\_align}}(i))$$
This design incentivizes the model to explore the ``cross-lingual optimal reasoning path'' while maintaining a robust defense against the structural instabilities inherent in tool-integrated RL.

\section{Experimental Setup}

\subsection{Datasets}
We utilize MKQA~\cite{longpre-etal-2021-mkqa} and XOR-TyDi QA ~\cite{asai-etal-2021-xor} datasets for evaluation. 
MKQA consists of 10k queries based on the NQ dataset~\cite{kwiatkowski-etal-2019-natural} translated into non-English languages, which support parallel languages in queries and knowledge.
XOR-TyDi QA comprises 40K information-seeking questions in 7 languages.
As shown in Table \ref{tab:lang_partition}, to evaluate the language transferability for the post-training methods, we define seen and unseen languages, where seen languages are used during RL training and unseen languages are only used at evaluation time.

In this paper, we follow~\citet{chirkova2024retrieval} and select a subset of queries encompassing 13 languages to maintain the diversity, including both Latin and non-Latin scripts. 
The details of statistics are provided in Appendix~\ref{app:dataset}.
\begin{table}[h]
\centering

\small
\begin{tabularx}{\linewidth}{l | l X}
\toprule
\textbf{Dataset} & \textbf{Type} & \textbf{Languages} \\
\midrule
\multirow{2}{*}{\textbf{MKQA}} & Training & \textproc{En}, \textproc{Ar}, \textproc{Fi}, \textproc{Fr}, \textproc{Ja}, \textproc{It}, \textproc{Ru}, \textproc{Zh} (8) \\
& Zero-shot & \textproc{De}, \textproc{Ko}, \textproc{Es}, \textproc{Pt}, \textproc{Th} (5) \\
\midrule
\multirow{2}{*}{\textbf{XOR-TyDi}} & Training & \textproc{En}, \textproc{Fi}, \textproc{Ja}, \textproc{Ko}, \textproc{Ru}, \textproc{Ar} (6) \\
& Zero-shot & \textproc{Bn}, \textproc{Te} (2) \\
\bottomrule
\end{tabularx}
\caption{Language partition for training and zero-shot evaluation.}
\label{tab:lang_partition}
\end{table}

\subsection{Implementation Details}
\paragraph{Base Models.} We select three LLMs with different sizes as backbones to implement ours: \texttt{\textbf{Qwen2.5-3B-Instruct}}, \texttt{\textbf{Qwen3-4B}}, and \texttt{\textbf{Qwen3-8B}}~\cite{yang2025qwen3}.
For the multilingual retriever, we use multilingual E5~\cite{wang2024multilingual} and Wikimedia\_dump\footnote{\url{https://huggingface.co/datasets/wikimedia/wikipedia}} as the knowledge base for all experiments.
To account for variability, the reported performance is averaged over three independent inference runs.
More details are in provided in Appendix~\ref{app:retriever} 
\paragraph{Evaluation Metrics.}
We follow the evaluation settings of~\citet{chirkova2024retrieval}, which consist of flexible exact match~(fEM) and recall on character 3-gram level~(c3Recall)~\cite{schick2023toolformer}. C3Recall measures the overlap between generated and gold answers at the character 3-gram level, providing a soft matching signal for multilingual evaluation.
Furthermore, we measure the correct language rate~(CLR), which is the percentage of outputs written in the user language by the models. CLR reflects the usability of multilingual RAG systems by assessing whether the model maintains the intended response language and avoids undesired language switching.

\subsection{Baselines}
To evaluate the effectiveness of \ours, we select the following baselines: (1) Direct inference without Retrieval. (2) Inference with Retrieval: native RAG~\cite{lewis2020retrieval}, IRCoT~\cite{trivedi2023interleaving}, and Search-o1~\cite{li2025search}. (3) Fine-Tuning Methods: Supervised fine-tuning (SFT)~\cite{chung2024scaling}, Argumentative Explanations~(D-RAG)~\cite{ranaldi-etal-2025-improving-multilingual}, and RL-based fine-tuning (Search-R1\footnote{we reproduce the Search-R1 method in the multilingual setting})~\cite{jin2025searchr1trainingllmsreason}. 
These baselines cover a broad spectrum of training-free and post-training RAG approaches, allowing for a comprehensive assessment of \ours. 
To make a fair comparison between different methods, we use the same settings of retriever, knowledge corpus, training data, and base LLMs. 
More details can be found in Appendix~\ref{app:baseline}.

\begin{table}[!t]
    \centering
    \definecolor{myblue}{RGB}{226,243,246}
    \definecolor{myred}{RGB}{238,228,241}
    \resizebox{0.48\textwidth}{!}{
    \begin{tabular}{lccccccc}
        \toprule
        \multirow{2}{*}{\textbf{Methods}}   & \multicolumn{3}{c}{\textbf{MKQA}} & \multicolumn{3}{c}{\textbf{XOR-Tydi QA}} \\
        \cmidrule(lr){2-4}\cmidrule(lr){5-7}
         \textbf{ } & \textbf{fEM} & \textbf{c3Recall} & \textbf{CLR} & \textbf{fEM} & \textbf{c3Recall} & \textbf{CLR} \\
        \midrule
\textbf{\texttt{Qwen2.5-3B-Instruct}} \\
        \rowcolor{myred}Direct Inference & 19.7 & 27.4 & 72.3 & 10.1 & 19.0 & 70.6 \\
        \rowcolor{myred}IRCoT & 22.4 & 34.9 & 65.3 & 18.5 & 27.9 & 69.4 \\
        \rowcolor{myred}Search-o1 & 27.7 & 35.0 & 70.2 & 21.1 & 30.7 & 67.9  \\
        \rowcolor{myred}RAG & 28.2 & 34.6 & 66.3 & 20.0 & 29.1 & 65.2  \\
        \rowcolor{myblue}D-RAG & 37.4 & 43.3 & 90.2 & 31.5 & 38.9 & 87.1  \\
        \rowcolor{myblue}SFT & 35.0 & 43.0 & 92.9 & 28.3 & 35.5 & 87.7 \\
        \rowcolor{myblue}Search-R1 & 22.6 & 34.8 & 83.6 & 18.4 & 32.0 & 86.3  \\
        \rowcolor{myblue}mSearch-R1 & 37.9 & 53.2 & 95.6 & 21.2 & 35.8 & 94.1  \\
        \rowcolor{red!20}\ours~(Ours) & \textbf{41.2$^*$} & \textbf{57.0$^*$} & \textbf{99.1$^*$} & \textbf{31.7$^*$}& \textbf{43.9$^*$} & \textbf{99.2$^*$}  \\
        \midrule
\textbf{\texttt{Qwen3-4B}} \\

        \rowcolor{myred}Direct Inference & 20.3 & 29.3 & 65.4 & 10.0 & 20.3 & 66.1 \\
        \rowcolor{myred}IRCoT & 20.3 & 28.8 & 68.2 & 20.7 & 28.1 & 69.6 \\
        \rowcolor{myred}Search-o1 & 22.1 & 28.6 & 72.4 & 19.6 & 26.7 & 73.1  \\
        \rowcolor{myred}RAG & 22.9 & 29.7 & 71.0 & 20.7 & 28.1 & 71.7  \\
        \rowcolor{myblue}D-RAG & 33.0 & 39.7 & 92.7 & 31.1 & 38.7 &  90.3 \\
        \rowcolor{myblue}SFT & 21.7 & 30.6 & 92.1 & 28.2 & 35.6 & 91.6 \\
        \rowcolor{myblue}Search-R1 & 24.3 & 34.7 & 85.8 & 17.8 & 29.8 & 87.2  \\
        \rowcolor{myblue}mSearch-R1 & 30.3 & 40.7 & 95.0 & 23.4 & 36.6 & 94.6  \\
        \rowcolor{red!20}\ours~(Ours) & \textbf{40.6$^*$} & \textbf{54.9$^*$} &  \textbf{98.9$^*$} & \textbf{32.4$^*$} & \textbf{45.7$^*$} & \textbf{99.4$^*$}  \\
        \midrule
\textbf{\texttt{Qwen3-8B}} \\

        \rowcolor{myred}Direct Inference & 26.2 & 24.3 & 66.8 & 12.3 & 22.7 & 67.5 \\
        \rowcolor{myred}IRCoT & 28.7 & 35.6 & 70.3 & 24.0 & 33.0 & 71.9 \\
        \rowcolor{myred}Search-o1 & 28.6 & 35.8 & 73.8 & 25.2 & 34.3 & 74.6  \\
        \rowcolor{myred}RAG & 27.6 & 34.8 & 74.2 & 22.7 & 31.8 & 74.9  \\
        \rowcolor{myblue}D-RAG & 38.0 & 46.7 & 94.1 & 33.5 & 42.3 & 93.5  \\
        \rowcolor{myblue}SFT & 29.8 & 38.3 & 93.4 & 30.0 & 38.9 & 94.0 \\
        \rowcolor{myblue}Search-R1 & 29.1 & 37.4 & 86.4 & 16.8 & 28.1 & 88.9  \\
        \rowcolor{myblue}mSearch-R1 & 40.6 & 50.8 & 94.8 & 27.2 & 38.5 & 94.2  \\
        \rowcolor{red!20}\ours~(Ours) & \textbf{47.6$^*$} & \textbf{63.2$^*$} & \textbf{99.4$^*$} & \textbf{38.3} & \textbf{48.9$^*$} & \textbf{99.3$^*$}  \\
        \bottomrule
    \end{tabular}%
     }
     \caption{The overall performance in all languages retrieval collections for MRAG. 
     The highest score based on each base model is highlighted in \textbf{bold}. 
     ``$*$'' denotes significant improvements with t-test at p < 0.01 between \ours~and mSearch-R1.
     }
    \label{tab:main_results}
\end{table}

\section{Results}

\subsection{Main Results}
\label{main_results}
\paragraph{Overall.}

As shown in Table~\ref{tab:main_results}, we investigate the performance of \ours~on two multilingual knowledge-intensive benchmarks.
The results demonstrate that our proposed method (\ours) outperforms several strong baselines in terms of all evaluation metrics for various LLMs.
In particular, post-training methods learn to contextualize cross-lingual reasoning processes to minimize their impact on the output language and demonstrate a clear advantage in correct language rate over training-free baselines, which our approach further enhances.
We further provide qualitative case studies in Appendix~\ref{appendix:case study} to illustrate how LcRL resolves knowledge bias and conflict in multilingual settings.

Although the multilingual search-augmented RL method~(mSearch-R1) facilitates the knowledge transfer in the multilingual settings compared to other post-training methods such as D-RAG and SFT, the parameters are still hard to optimize, and its performance has not been significantly improved.
The limitations arise from optimization instability and knowledge conflicts, in which deep, multi-turn retrieval over multilingual collections is incorporated into the search engine calling process. 
We address these issues by incorporating a language-coupled mechanism into the rollout and reward modules, enabling stable training and effective multilingual knowledge transfer within groups, which achieves significant performance and successfully transfers to unseen languages.
In addition, we also conduct experiments on unseen languages that were not included in the training data. \ours~also outperforms other methods. Results show that \ours~demonstrates superior generalization performance compared to all baseline methods. The results and further explanations for each language are provided in Appendix~\ref{app:allResults}.

\begin{figure}[!t]
    \centering
    \includegraphics[width=\linewidth]{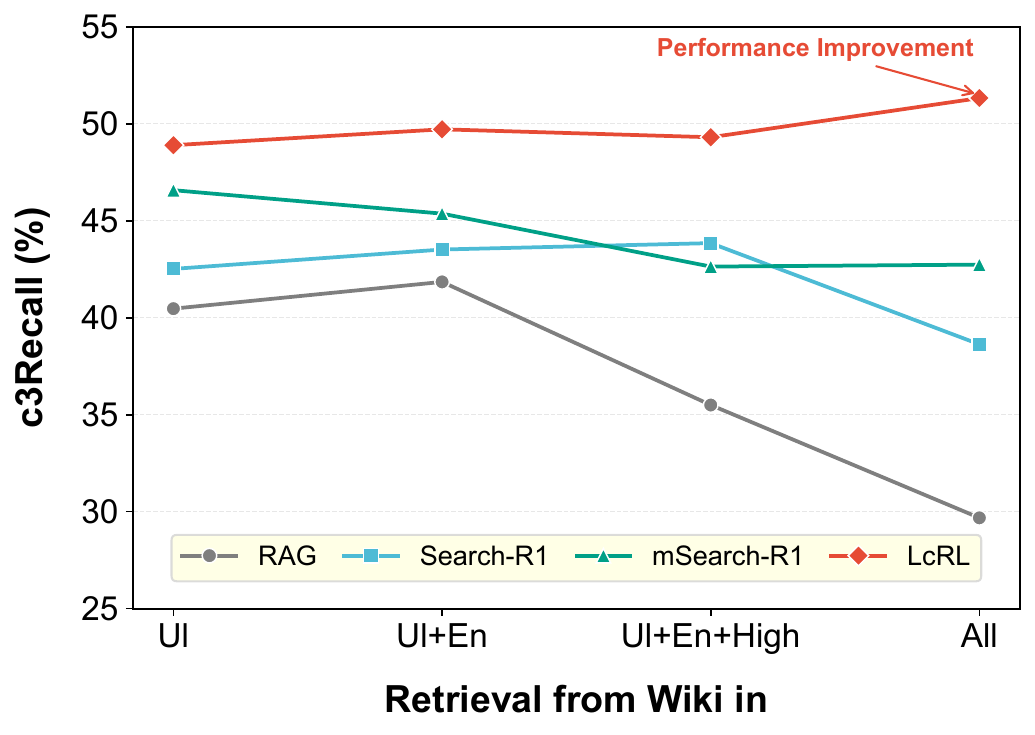}
    \caption{Performance with different retrieval options. Ul denotes the language of the user query, and High represents a randomly selected high-resource language.}
    \label{fig:count}
\end{figure}
\begin{table}[!t]
    \centering
    \small
    \begin{tabular}{lcccc}
        \toprule
         \textbf{ } & \textbf{fEM} & \textbf{c3Recall} & \textbf{CLR} \\
        \midrule
        \rowcolor{gray!20}Our Full Model & 41.2 & 57.0 & 99.1 \\
        \midrule
        w/o \textproc{Lc} Reward & 30.8 & 42.2 & 98.1 \\
        w/o c3Recall Reward & 18.0 & 20.2 & 97.8  \\
        w/o \textproc{Lc} Rollout & 30.4 & 45.7 & 98.7 \\
        w/o multi-language Rollout & 27.9 & 38.5 & 84.6 \\
        \midrule
        Replace by PPO & 15.5 & 21.7 & 72.3\\
        \bottomrule
    \end{tabular}%
     \caption{Results of \ours~with algorithms and components based on \texttt{\textbf{Qwen2.5-3B-Instruct}} for MKQA.}
    \label{tab:ablation_results}
\end{table}
\paragraph{Performance on Language Count in Multilingual Collections.}
As shown in Figure~\ref{fig:count}, we investigate the impact of increasing language diversity in the retrieval collections on different RAG methods. 
The results show that only our method exhibits a steady improvement, whereas all other approaches suffer a sharp performance degradation when the number of languages in the retrieval collections exceeds two. 
This degradation is attributable to knowledge conflicts arising from factual inconsistencies across languages in the collections. 
Our method explicitly addresses this issue through an anti-consistency penalty in the reward model. 
Moreover, detailed per-language results and analysis are provided in Appendix~\ref{app:allResults}, including performance comparisons across both seen and unseen languages, as well as additional analyses on data scaling and language coverage. The analysis shows that the language-coupled approach prevents negative transfer from low-resource languages where the multilingual curse is prevalent in high-resource languages.


\begin{figure}[!t]
    \centering
    \includegraphics[width=\linewidth]{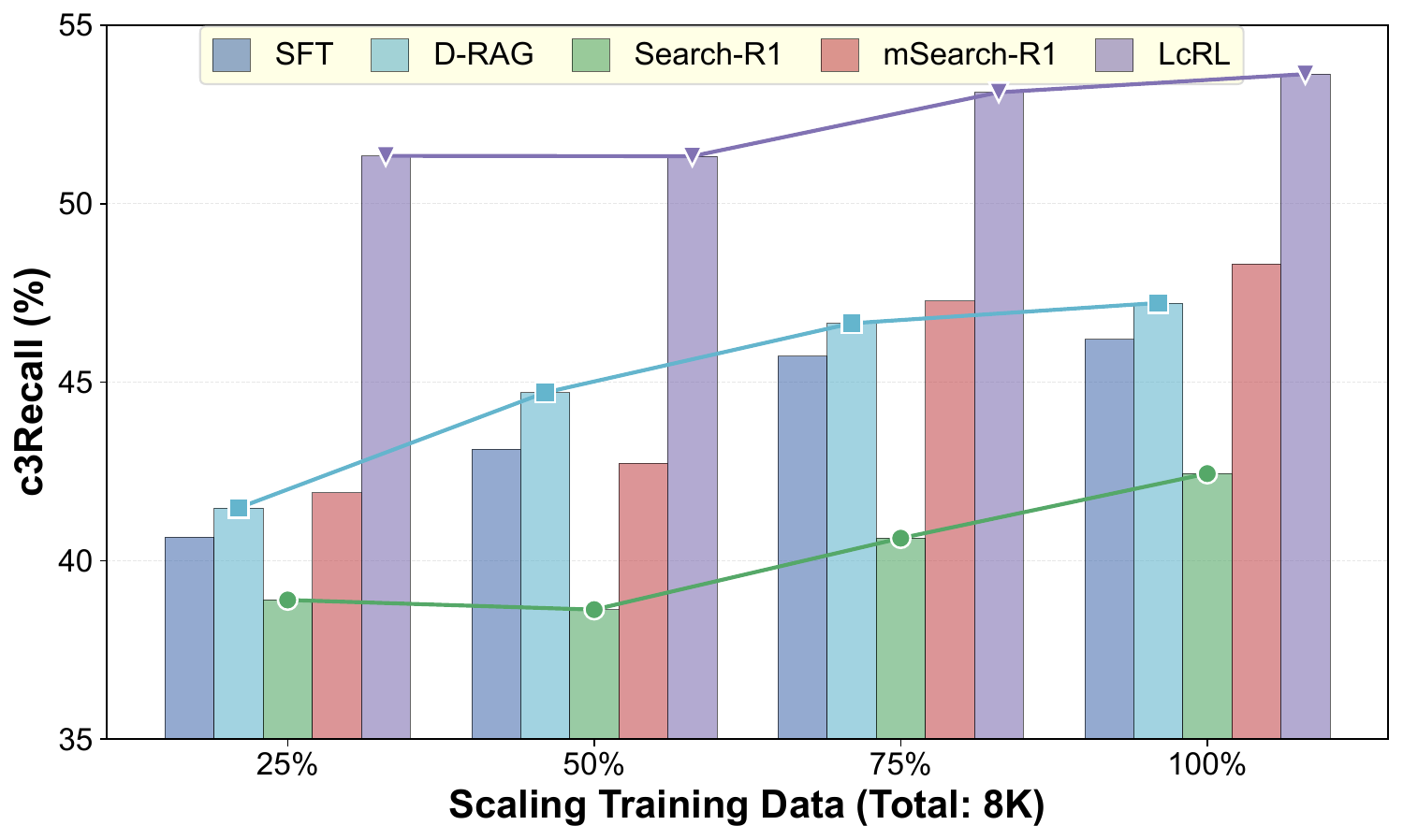}
    \caption{Results of post-training methods with scaling utilized data.}
    \label{fig:effect_sca}
\end{figure}
\begin{figure}[!t]
    \centering
    \includegraphics[width=\linewidth]{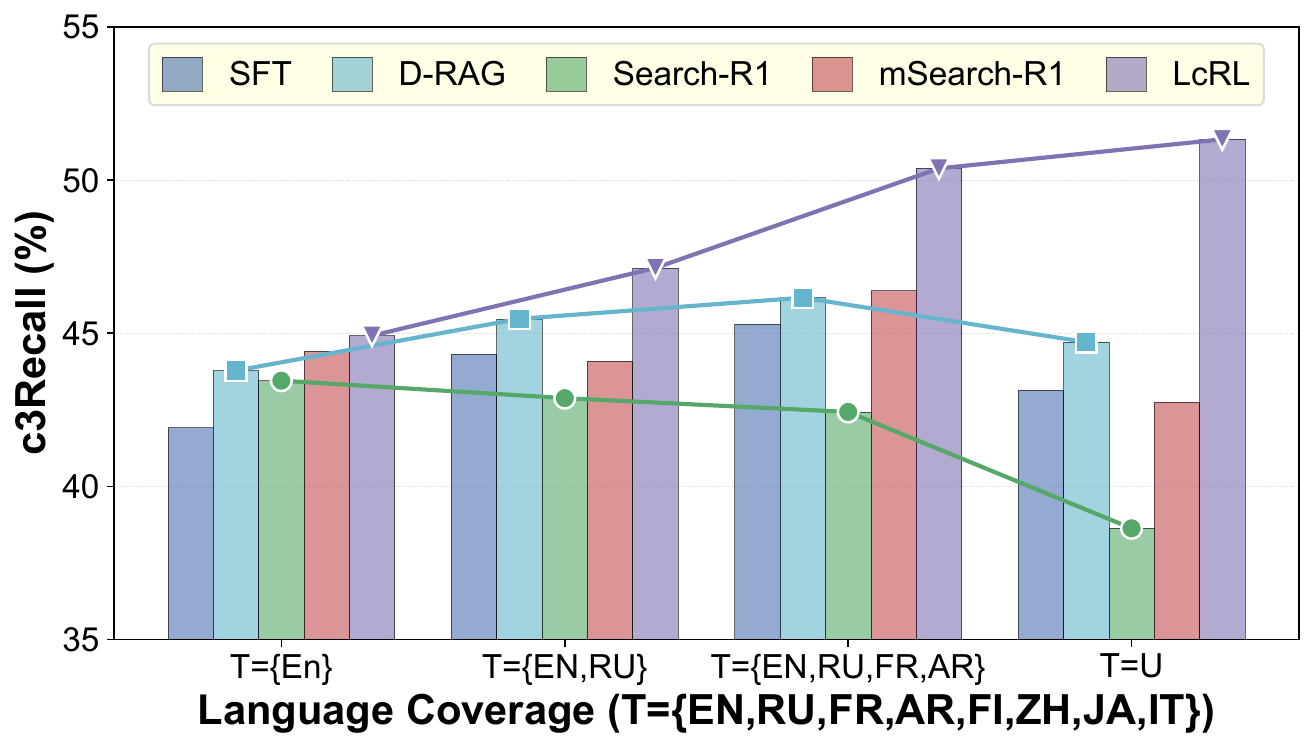}
    \caption{Performance of post-training methods with different language coverages.}
    \label{fig:effect_lc}
\end{figure}
\begin{figure*}[!t] 
    \centering
    \begin{subfigure}{0.45\linewidth}
        \centering
        \includegraphics[width=\linewidth]{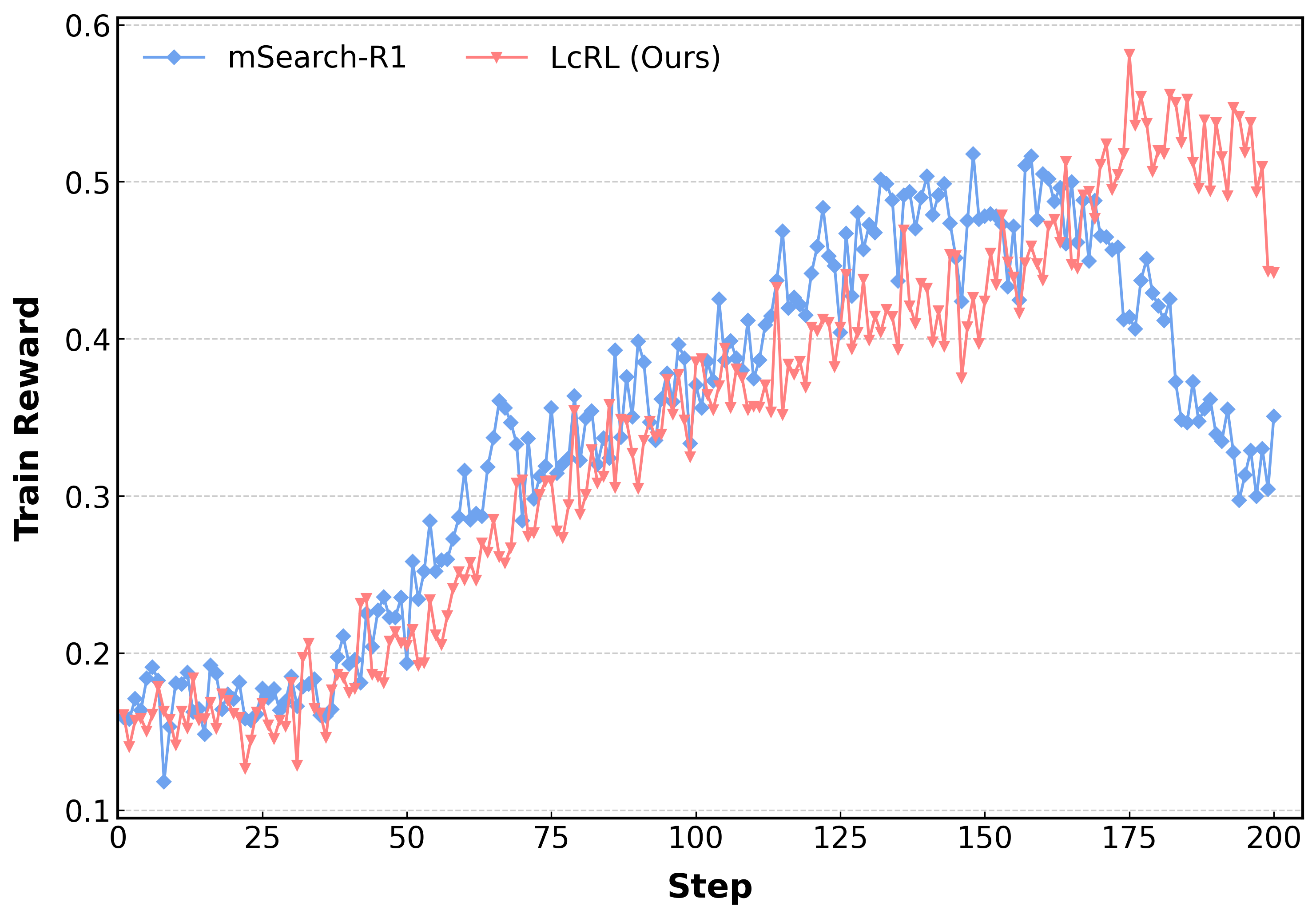} 
        \caption{Train reward.}
        \label{fig:sub1}
    \end{subfigure}
    \begin{subfigure}{0.45\linewidth}
        \centering
        \includegraphics[width=\linewidth]{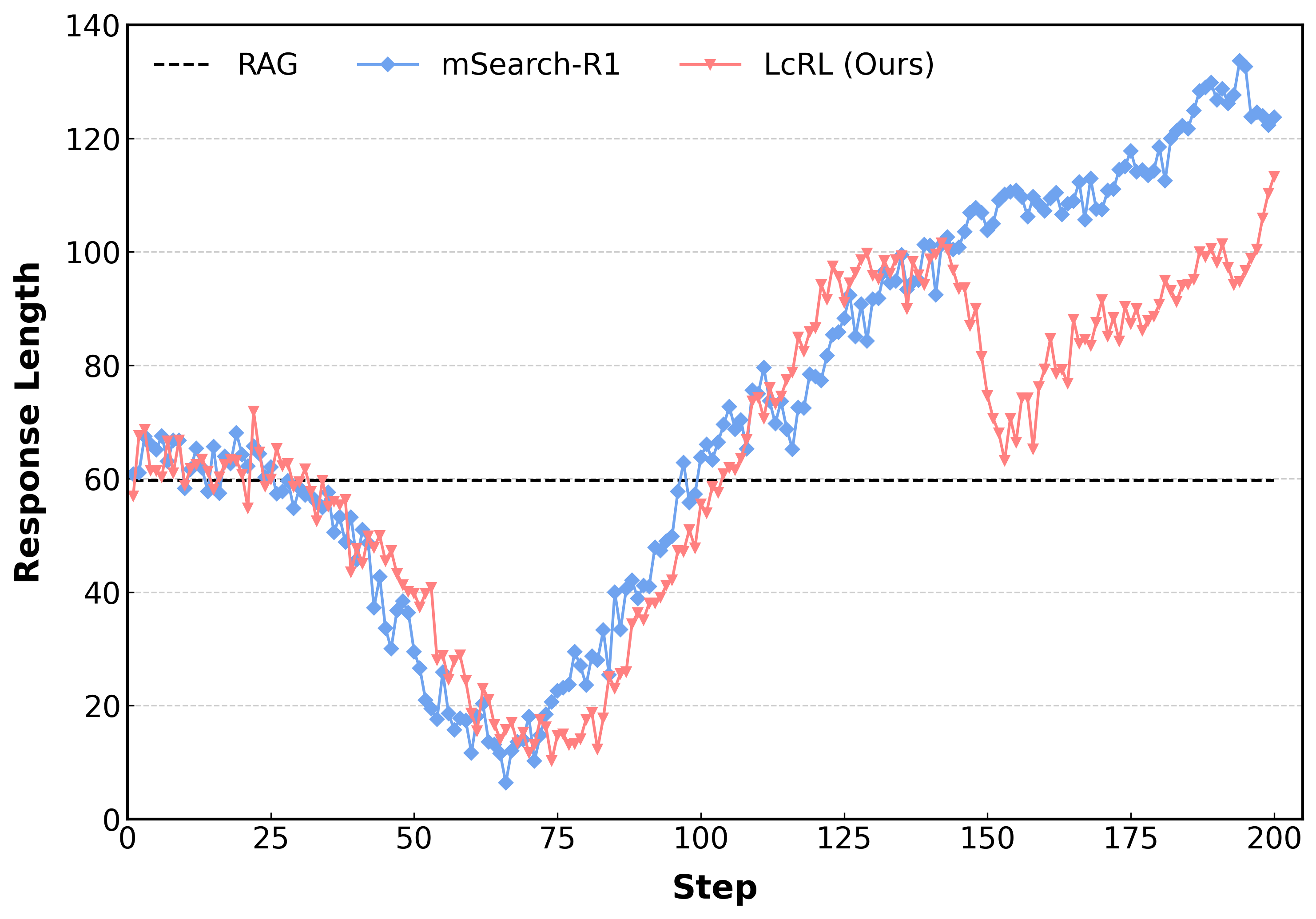} 
        \caption{Response length.}
        \label{fig:sub2}
    \end{subfigure}
    \caption{Comparisons of train reward and response length between mSearch-R1 and ours.}
    \label{fig:main}
\end{figure*}
\subsection{Ablation Studies}
As shown in Table~\ref{tab:ablation_results}, we further investigate the effectiveness of our method with different algorithms and components.
\textit{First}, we replace the outcome-based reward function. The EM objective provides a sparse optimization signal for MRAG. By introducing n-gram preference optimization, we observe a 22\% and a 12.8\% gain with respect to c3Recall and fEM, respectively. The proposed language-coupled mechanism delivers further performance gains by suppressing training collapse arising from language-consistency error modes.
\textit{Second}, we replace the search engine calling strategy in the rollout module. The results show that multi-turn retrieval not only improves overall performance but also reduces language error rates, and our proposed language-coupled multi-turn retrieval further strengthens the interaction between LLMs and the search engine in multilingual scenarios.
\textit{Finally}, we compare different RL algorithms (GRPO v.s. PPO). The results demonstrate the superiority of GRPO in multilingual settings. It is attributed to the group-based learning mechanism that facilitates cross-lingual generalization, and the limitations of value LLMs in multilingual training.

\section{Analysis}

\subsection{Effects of Training Data}
To further investigate the generalization and cross-lingual transferability of post-training methods, we systematically vary both the scale and the language coverage of the training corpus. 
Overall, our method exhibits positive robustness when handling a larger number of languages.

\paragraph{Data Scale.}
Figure~\ref{fig:effect_sca} presents a comparison of post-training methods under training corpus scaling. SFT-based methods perform poorly in low-resource settings but show substantial performance gains as the data scale increases. In contrast, RL-based methods achieve competitive performance even with limited data, with scaling providing only marginal improvements. However, once the training corpus reaches a sufficient size, RL-based methods experience significant performance gains.

\paragraph{Language Coverage.}
Figure~\ref{fig:effect_lc} shows that our method achieves superior performance when the training corpus covers a large number of languages, indicating that it effectively mitigates the ``multilingual curse'' and highlighting the distinct effectiveness in multilingual scenarios. Although performance decreases as the number of languages is reduced, our approach consistently outperforms competing methods. When the multilingual setting is absent (i.e., the training corpus contains only a single language), our method exhibits performance comparable to other approaches.

\subsection{Response Length and Training Stability}

As illustrated in Figure~\ref{fig:sub1}, we analyze the training dynamics to verify the stability of \ours, and mSearch-R1 exhibits a sharp ``reward collapse'' starting around step 160. In contrast, \ours~maintains a robust upward trajectory throughout the training process, stabilizing at higher reward levels without succumbing to performance degradation. This indicates that our language-coupled reward modeling effectively regularizes the policy against error mode collapse.
Furthermore, Figure~\ref{fig:sub2} highlights the efficiency of the generated reasoning trajectories. While mSearch-R1 suffers from uncontrolled length expansion, which is a common symptom of reasoning loops in RL. \ours~effectively regulates the generation process. In the later training stages, \ours~achieves a response length reduction of approximately 12\% compared to mSearch-R1, indicating that \ours~does not rely on redundant token generation to improve performance and optimize for more concise multilingual reasoning paths.

\section{Conclusion}
In this work, we propose a language-coupled GRPO method for MRAG, which leverages multilingual and multi-turn decision-making with an anti-consistency reward. 
It can adaptively determine the need for retrieving external knowledge and effectively harmonize contradictory knowledge obtained from different languages, mitigating the issue of knowledge bias and conflicts. 
Experimental results demonstrate that our method exhibits strong robustness under resource-constrained conditions and is effective in dealing with an increasing number of language counts in retrieval collections.

\section*{Limitation}
Although the effectiveness shown in our proposed method, some potential limitations should be addressed in future studies.

First, current evaluations typically reuse existing QA datasets combined with generic Wikipedia dumps.
Due to the lack of dedicated query-document relevance judgments annotations in multilingual RAG scenarios, we are unable to systematically evaluate the correlation between the search results evaluated by traditional IR metrics and the answer results produced by generator models. 
Second, due to cost constraints, we evaluate our method on three representative multilingual LLMs, which only scratches the surface of the world's vast array of open-resourced models.
Furthermore, as this study prioritizes the interaction between LLMs with multilingual retrieved evidence over the optimization of retrieval algorithms themselves, we utilize multilingual-e5-base as our fixed retriever. This model represents a widely adopted and computationally efficient baseline for multilingual dense retrieval. 
While a more expansive experimental configuration might yield higher absolute performance, the current setup provides a robust foundation for evaluating our core contributions.

\section*{Ethics Statement}
We did not explicitly address ethical considerations in this work. All data used in our experiments were obtained exclusively from open and publicly available benchmarks, as detailed in the appendices. Statistics related to language distributions in commonly used pre-training datasets were sourced from official open-resourced technical report. Our analysis and methodology were conducted with sensitivity, and do not involve or affect attributes related to gender, sex, or race.

\bibliography{custom}

\clearpage
\appendix
\section*{Appendix}
\section{Details in Experiments}

\subsection{Datasets}
\label{app:dataset}
The details of training and evaluation data are presented in Table~\ref{tab:lang_partition} and~\ref{tab:data_stats}. For MKQA, we initially select a subset of 2k instances per language for the test set. 
For XOR-TyDi QA, we select the already segmented validation set in 8 languages and ground questions in Wikipedia in the same language as the question.

\begin{table}[h]
\centering

\resizebox{0.48\textwidth}{!}{
\begin{tabular}{l  ccc  ccc}
\toprule
\multirow{2}{*}{\textbf{Dataset}} & \multicolumn{3}{c}{\textbf{Training Set}} & \multicolumn{3}{c}{\textbf{Test Set}} \\
\cmidrule(lr){2-7}
& \# Lang & \# / Lang & \textbf{Total} & \# Lang & \# / Lang & \textbf{Total} \\
\midrule
MKQA & 8 & 1,000 & 8,000 & 13 & 2000 & 16,601 \\
XOR-TyDi & 6 & 1,000 & 6,000 & 7 & \textit{U*} & 2,560 \\
\bottomrule
\end{tabular}}

\vspace{2pt}
\begin{flushleft}
\scriptsize
\textbf{*Note:} XOR-TyDi counts in group \textit{U}: fi(615), ja(433), ko(371), ru(568), ar(708), bn(427), te(351).
\end{flushleft}
\caption{Number (\#) of training and evaluation instances.}
\label{tab:data_stats}
\end{table}

\subsection{Retriever and Generator}
\label{app:retriever}

\paragraph{Retriever.} We pick a strong and publicly available Multilingual-E5-base\footnote{\url{https://huggingface.co/intfloat/multilingual-e5-base}} as our multilingual retriever, and Wikimedia\_dump as the knowledge base for all experiments. We split the Wikipedia article into chunks of 100 words (or
100 Unicode characters for non-whitespace separated
languages) and prepending the article title to each chunk.
Following Search-R1~\cite{jin2025searchr1trainingllmsreason}, we set the number of retrieved passages to 3 across all retrieval-based methods.

\paragraph{Generator.} We use various strong LLM, including recently released Qwen3-4B\footnote{\url{https://huggingface.co/Qwen/Qwen3-4B}} and Qwen3-8B\footnote{\url{https://huggingface.co/Qwen/Qwen3-8B}}, as well as widely-used LLM Qwen2.5-3B-Instruct\footnote{\url{https://huggingface.co/Qwen/Qwen2.5-3B-Instruct}}. 

\begin{table}[!t]
    \centering
    \small
    \definecolor{colorMethods}{RGB}{255, 255, 255}
    \definecolor{colorGroup1}{RGB}{226, 243, 246}
    \definecolor{colorGroup2}{RGB}{238, 228, 241}
    \definecolor{colorAve}{RGB}{250, 235, 215}

    \resizebox{0.48\textwidth}{!}{ 
    \begin{tabular}{ 
        >{\columncolor{colorMethods}}l       
        >{\columncolor{colorGroup1}}c        
        *{4}{>{\columncolor{colorGroup1}}c}  
        >{\columncolor{colorGroup2}}c        
        >{\columncolor{colorGroup2}}c        
        >{\columncolor{colorAve}}c           
    }
        \toprule
        \rowcolor{colorMethods}
        \multirow{2}{*}{\textbf{Methods}}   & \multicolumn{7}{c}{\textbf{Language}} & \\
        \cmidrule(lr){2-8}
        \rowcolor{colorMethods}
        \textbf{ } 
        & \textbf{\textproc{Ar}} 
        & \textbf{\textproc{Fi}} 
        & \textbf{\textproc{Ja}} 
        & \textbf{\textproc{Ko}} 
        & \textbf{\textproc{Ru}} 
        & \textbf{\textproc{Bn}} 
        & \textbf{\textproc{Te}}  
        & \textbf{\textproc{Ave}} \\
        \midrule
        
        \multicolumn{9}{c}{\cellcolor{colorMethods}\textbf{\texttt{Qwen2.5-3B-Instruct}}} \\
        \midrule
        Direct Inference & 26.9 & 18.8 & 11.1 & 17.4 & 21.0 & 10.2 & \underline{7.1} & 16.1 \\
        IRCoT  & 39.4 & 35.7 & 22.7 & 9.6 &32.3 & 9.3 & 7.0 & 22.3  \\
        Search-o1  & 42.5 & 40.4 & 21.8 & 12.5 & 36.3 & 9.7 & 6.1 & 24.2  \\
        RAG  & 40.1 & 39.1 & 21.2 & 10.8 & 34.3 & 10.2 & 6.0 & 23.1 \\
        D-RAG  & \underline{50.4} & \underline{48.7} & 31.9 & 20.6 & 43.1 & 9.7 & 6.7 & \underline{30.1} \\
        SFT  & 48.0 & 45.5 & \underline{32.2} & 12.4 & 39.3 & 5.7 & 6.1 & 27.0 \\
        Search-R1  & 37.5 & 41.5 & 20.7 & 21.7 & 38.7 & \textbf{12.7} & 6.5 & 25.6  \\
        mSearch-R1  & 44.7  & 45.6 & 21.1 & \underline{23.7} & \underline{44.0} &  \underline{12.5} & 7.0 & 28.4 \\

        \ours~(Ours)  & \textbf{53.4} & \textbf{57.6} & \textbf{32.4} & \textbf{30.3} & \textbf{45.8} & 11.5 & \textbf{8.6} & \textbf{34.2}  \\
        \midrule
        
        \multicolumn{9}{c}{\cellcolor{colorMethods}\textbf{\texttt{Qwen3-4B}}} \\
        \midrule
        Direct Inference & 27.1 & 22.6 & 12.3 & 17.7 & 21.7 & \textbf{12.9} & \underline{11.2} & 17.9 \\
        IRCoT  & 31.9 & 43.9 & 18.3 & 13.0 & 26.6 & 5.4 & 6.8 & 20.9  \\
        Search-o1  & 34.2 & 48.2 & 18.8 & 13.4 & 27.7 & 5.0 & 9.8 & 22.4  \\
        RAG  & 32.7 & 46.1 & 20.2 & 12.4 & 29.0 & 4.8 & 7.8 & 21.8 \\
        D-RAG  & \underline{44.8} & \underline{56.2} & \underline{33.6} & 17.3 & \underline{41.6} & 9.8 & 10.5 & \underline{30.5} \\
        SFT  & 41.4 & 55.0 & 29.7 & 14.1 & 38.0 & 5.9 & 8.7 & 27.5 \\
        Search-R1  & 39.6 & 36.6 & 19.8 & 24.4 & 28.7 & 8.9 & 4.5 & 23.2  \\
        mSearch-R1  & 44.4 & 40.9 & 25.9 & \textbf{32.6} & 39.3 & 10.0 & 6.1 & 28.4  \\
        \ours~(Ours)  & \textbf{56.1} & \textbf{59.5} & \textbf{34.2} & \underline{29.9} & \textbf{48.5} & \underline{11.5} & \textbf{11.4} & \textbf{35.9}  \\
        \midrule
        
        \multicolumn{9}{c}{\cellcolor{colorMethods}\textbf{\texttt{Qwen3-8B}}} \\
        \midrule
        Direct Inference & 29.1 & 27.0 & 13.1 & 21.3 & 22.9 & \textbf{17.8} & \textbf{16.7} & 21.1  \\
        IRCoT  & 42.7 & 47.2 & 22.4 & 15.2 & 37.6 & 13.1 & 12.0 & 27.2  \\
        Search-o1  & 43.6 & 47.3 & 25.6 & 14.2 & 40.7 & 13.4 & 10.4 & 27.9  \\
        RAG  & 41.1 & 44.1 & 23.8 & 13.2 & 36.6 & 13.9 & 11.3 & 26.3 \\
        D-RAG  & \underline{52.3} & \underline{53.9} & \underline{36.7} & 22.8 & \underline{45.9} & 13.3 & \underline{12.6} & \underline{33.9} \\
        SFT  & 49.7 & 50.0 & 33.3 & 19.0 & 42.7 & 13.2 & 10.9 & 31.3 \\
        Search-R1  & 38.9 & 32.2 & 20.7 & 19.1 & 29.7 & 9.6 & 5.0 & 22.2  \\
        mSearch-R1  & 46.0 & 45.4 & 26.0 & \underline{33.1} & 43.2 & 12.9 & 8.0 & 30.5  \\
        \ours~(Ours)  & \textbf{59.1} & \textbf{63.1} & \textbf{39.0} & \textbf{35.0} & \textbf{48.2} & \underline{15.8} & 11.0 & \textbf{38.8}  \\      
        \bottomrule
    \end{tabular}
    }
    \caption{Main results~(\%) of XOR-Tydi QA. Background colors distinguish between \colorbox{colorGroup1}{seen languages} and \colorbox{colorGroup2}{unseen languages}. The highest score on each base model is highlighted in \textbf{bold}. The second highest score on each base model is highlighted in \underline{underline}.}
    \label{tab:detail_xor}
\end{table}

\begin{table*}[!t]
    \centering
    \small
    \setlength{\tabcolsep}{5pt} 
    \definecolor{colorMethods}{RGB}{255, 255, 255}
    \definecolor{colorGroup1}{RGB}{226, 243, 246}
    \definecolor{colorGroup2}{RGB}{238, 228, 241}
    \definecolor{colorAve}{RGB}{250, 235, 215}

    \begin{tabular}{ 
        >{\columncolor{colorMethods}}l      
        *{8}{>{\columncolor{colorGroup1}}c} 
        *{5}{>{\columncolor{colorGroup2}}c} 
        >{\columncolor{colorAve}}c          
    }
        \toprule
        \rowcolor{colorMethods}
        \multirow{2}{*}{\textbf{Methods}}   & \multicolumn{13}{c}{\textbf{Language}} & \\
        \cmidrule(lr){2-14}
         \rowcolor{colorMethods} 
        \textbf{ } & \textbf{\textproc{En}} & \textbf{\textproc{Fr}} & \textbf{\textproc{It}} & \textbf{\textproc{Ru}} & \textbf{\textproc{Fi}} & \textbf{\textproc{Zh}} & \textbf{\textproc{Ja}} & \textbf{\textproc{Ar}} & \textbf{\textproc{Es}} & \textbf{\textproc{Pt}} & \textbf{\textproc{Ko}} & \textbf{\textproc{Th}} & \textbf{\textproc{De}}  & \textbf{\textproc{Ave}} \\
        \midrule
        
        \multicolumn{15}{c}{\cellcolor{colorMethods}\textbf{\texttt{Qwen2.5-3B-Instruct}}} \\
        \midrule
        Direct Inference & 51.8 & 39.8 & 36.9 & 21.5 & 23.6 & 13.2 & 14.9 & 17.5  & 39.3 & 37.8 & 8.8 & 18.0 & 39.7 & 27.9  \\
        IRCoT & 61.8 & 50.1 & 45.6 & 33.9 & 31.7 & 15.6 & 17.1 & 23.3 & 57.3 & 55.3 & 12.7 & 20.0 & 48.1 & 36.3 \\
        Search-o1 & 60.9 & 43.2 & 50.8 & 27.4 & 37.6 & 18.9 & 25.1 & 16.3 & 58.7 & 58.9 & 14.2 & 22.6 & 57.5 & 37.9 \\
        RAG & 64.1 & 46.4 & 46.4 & 30.7 & 33.1 & 14.7 & 20.6 & 20.9 & 54.9 & 52.2 & 10.6 & 18.8 & 51.9 & 35.8  \\
        D-RAG & 72.6 & 54.9 & 55.1 & 38.8 & 42.3 & 22.9 & 29.4 & 30.6 & 62.8 & 59.7 & \underline{18.9} & 26.1 & 61.3 & 44.3 \\
        SFT & 60.1 & 55.8 & 52.8 & 44.9 & 41.0 & 23.3 & \textbf{33.5} & 32.6 & 56.9 & 53.4 & 16.5 & 28.0 & 54.7 & 36.3 \\
        Search-R1 & 59.7 & 45.9 & 47.9 & 36.9 & 36.5 & 16.4 & 15.2 & 20.2 & 50.3 & 43.2 & 12.2 & 19.8 & 44.8 & 34.5 \\
        mSearch-R1 & \underline{74.6} & \underline{73.9} & \underline{72.5} & \underline{53.6} & \underline{63.5} & \underline{26.6} & 28.2 & \underline{33.0} & \underline{67.4} & \underline{72.8} & 18.4 & \underline{32.9} & \underline{73.3} & \underline{53.1} \\
        \ours~(Ours) & \textbf{81.9} & \textbf{75.6} & \textbf{76.0} & \textbf{57.6} & \textbf{68.7} & \textbf{27.8} & \underline{30.8} & \textbf{37.9} & \textbf{78.2} & \textbf{74.4} & \textbf{25.0} & \textbf{40.3} & \textbf{74.9} & \textbf{57.6} \\
        \midrule
        
        \multicolumn{15}{c}{\cellcolor{colorMethods}\textbf{\texttt{Qwen3-4B}}} \\
        \midrule
        Direct Inference & 51.9 & 37.8 & 38.6 & 26.4 & 24.6 & 18.9 & 16.4 & 20.1 & 39.4 & \underline{39.2} & 10.2 & 19.5 & 38.6 & 29.4 \\
        IRCoT & 41.9 & 36.4 & 24.7 & 33.3 & 34.2 & 22.7 & 17.4 & 20.0 & 40.3 & 30.0 & 9.8 & 20.9 & 47.5 & 29.1 \\
        Search-o1 & 43.9 & 34.1 & 25.4 & 31.0 & 37.3 & 21.0 & 18.0 & 17.8 & 42.1 & 27.7 & 11.0 & 18.4 & 48.4 & 28.9 \\
        RAG & 46.6 & 32.9 & 29.9 & 29.9 & 39.9 & 19.3 & 22.5 & 16.6 & 45.9 & 26.1 & 13.7 & 17.1 & 51.6 & 30.1 \\
        D-RAG & 62.7 & 48.2 & 45.6 & 33.9 & \underline{45.4} & 24.9 & \underline{27.2} & \underline{29.4} & \underline{52.2} & 31.7 & 18.9 & 23.5 & \underline{58.0} & \underline{38.6} \\
        SFT & 53.2 & 38.6 & 39.4 & 28.4 & 26.1 & 19.2 & 17.5 & 22.0 & 48.1 & 32.7 & 19.9 & 24.0 & 54.4 & 32.6 \\
        Search-R1 & 58.5 & 47.5 & 46.1 & 31.7 & 32.5 & 21.6 & 19.3 & 20.5 & 33.8 & 33.0 & 13.9 & 20.5 & 28.5 & 31.3 \\
        mSearch-R1  & \underline{64.7} & \underline{53.2} & \underline{52.8} & \underline{38.1} & 39.2 & \underline{27.0} & 24.9 & 25.9 & 40.4 & 38.4 & \underline{20.1} & \underline{26.9} & 34.9 & 37.4 \\
        \ours~(Ours) & \textbf{73.1} & \textbf{60.9} & \textbf{73.5} & \textbf{59.5} & \textbf{69.1} & \textbf{29.9} & \textbf{33.4} & \textbf{39.8} & \textbf{64.9} & \textbf{54.5} & \textbf{21.6} & \textbf{29.2} & \textbf{60.0} & \textbf{51.5} \\
        \midrule
        
        \multicolumn{15}{c}{\cellcolor{colorMethods}\textbf{\texttt{Qwen3-8B}}} \\
        \midrule
        Direct Inference & 59.5 & 46.5 & 45.4 & 29.0 & 34.3 & 21.6 & 18.7 & 19.4 & 47.9 & 44.9 & 12.5 & 22.3 & 46.6 & 34.5 \\
        IRCoT & 54.2 & 38.5 & 36.9 & 33.1 & 46.0 & 22.8 & 25.9 & 27.4 & 58.7 & 29.7 & 18.9 & 17.8 & 52.1 & 35.5 \\
        Search-o1 & 53.3 & 40.7 & 35.6 & 35.3 & 44.9 & 22.2 & 28.3 & 26.1 & 57.5 & 31.7 & 18.5 & 20.4 & 51.6 & 35.9 \\
        RAG & 52.1 & 39.9 & 34.1 & 34.7 & 43.8 & 21.3 & 27.0 & 25.4 & 56.5 & 30.9 & 17.1 & 19.8 & 50.4 & 34.8 \\
        D-RAG & 68.8 & 56.4 & 53.6 & 44.1 & 53.0 & \underline{28.3} & \underline{35.1} & \underline{34.6} & 57.4 & 40.7 & 19.1 & 24.5 & 59.3 & 44.2 \\
        SFT & 55.4 & 42.9 & 36.0 & 38.9 & 47.1 & 25.6 & 30.5 & 29.7 & 60.9 & 34.6 & 19.7 & 22.4 & 55.3 & 38.4 \\
        Search-R1 & 56.7 & 42.3 & 33.7 & 39.9 & 45.7 & 22.5 & 31.2 & 27.6 & 52.4 & 33.4 & 15.8 & 20.5 & 52.3 & 36.5 \\
        mSearch-R1  & \underline{71.0} & \underline{70.5} & \underline{69.4} & \underline{47.5} & \underline{59.8} & 23.8 & 31.3 & 33.3 & \underline{74.7} & \underline{73.2} & \underline{20.1} & \underline{33.3} & \underline{72.5} & \underline{52.3} \\
        \ours~(Ours) & \textbf{88.3} & \textbf{78.9} & \textbf{78.2} & \textbf{63.5} & \textbf{75.0} & \textbf{35.4} & \textbf{39.3} & \textbf{46.9} & \textbf{84.5} & \textbf{80.0} & \textbf{33.8} & \textbf{48.6} & \textbf{80.6} & \textbf{64.1} \\
        \bottomrule
    \end{tabular}
     \caption{Main results~(\%) of MKQA. Background colors distinguish between \colorbox{colorGroup1}{seen languages} and \colorbox{colorGroup2}{unseen languages}. The highest score on each base model is highlighted in \textbf{bold}. The second-highest score on each base model is highlighted in \underline{underline}.}
    \label{tab:detail_mkqa}
\end{table*}

Training is performed on a single node with 4 A800 GPUs. We use a total batch size of 512, with a mini-batch size of 256 and a micro-batch size of 16. For GRPO training, we set the policy LLM learning rate to 1e-6 and sample 3 responses per prompt. For our auxiliary anti-consistency penalty introduced in section 3.3, $\lambda$ is set to 0.02.

The group size $G$ is treated as a hyperparameter and is not necessarily equal to the number of available languages $N$. When $G < N$, we always include the original-language query and randomly sample the remaining $G-1$ queries from other seen languages. We search $G$ in the range of 3 to 8 and select $G=5$, which is consistent with the optimal setting reported in Search-R1.

\subsection{Baselines}
\label{app:baseline}

We provide a brief introduction to the baseline methods used in our experiments.
\begin{itemize}
    \item Direct Inference. The model generates answers directly based on its internal parametric knowledge without any external retrieval. 
    \item RAG~\cite{lewis2020retrieval}. The traditional Retrieval-Augmented Generation framework. Documents are retrieved in a single step to provide context for the final generation.
    \item IRCoT~\cite{trivedi2023interleaving}. Interleaving Retrieval with Chain-of-Thought Reasoning. It guides the retrieval process by using the intermediate steps of a CoT as queries.
    \item Search-o1~\cite{li2025search}. An agentic search-enhanced framework that allows the model to autonomously plan, execute search queries, and refine its internal thought process before producing the final answer.
    \item D-RAG~\cite{ranaldi-etal-2025-improving-multilingual}. A dialectic reasoning argumentation approach that systematically evaluates retrieved information by comparing, contrasting, and resolving conflicting perspectives.
    \item SFT~\cite{chung2024scaling}. Supervised Fine-Tuning. LLMs are trained by queries and retrieved documents.
    \item Search-R1~\cite{jin2025searchr1trainingllmsreason}. A framework that trains LLMs to reason with search through Reinforcement Learning (RL). We also extend the Search-R1 to the Multilingual Search-augmented RL method~(mSearch-R1) with just Multi-Turn and Multilingual Retrieval introduced in section~\ref{sec:multi-turn}.

\end{itemize}

\begin{table*}[!t]
\centering

\begin{tcolorbox}[colback=bglight, colframe=gray!50, arc=4pt, boxrule=0.5pt, left=10pt, right=10pt, top=10pt, bottom=10pt]
\small
\textbf{System Prompt:} \\
Answer the given question. You must conduct reasoning inside {\color{tagblue}\texttt{<think>}} and {\color{tagblue}\texttt{</think>}} first every time you get new information. \\
After reasoning, if you find you lack some knowledge, you can call a search engine by {\color{tagpurple}\texttt{<search>}} query {\color{tagpurple}\texttt{</search>}}, and it will return the top searched results between {\color{taggreen}\texttt{<information>}} and {\color{taggreen}\texttt{</information>}}. \\

\textbf{Hierarchical Search Logic:} \\
You can search at least 2 times: \\
1. First search in \textbf{\{lang\_name\}} knowledge base. \\
2. Second search in \textbf{English and other languages} (e.g., French) to provide comprehensive information and resolve potential conflicts.

If needed, continue searching in English. \\

\textbf{Output Constraints:} \\
If no further external knowledge is needed, provide the answer inside {\color{tagorange}\texttt{<answer>}} and {\color{tagorange}\texttt{</answer>}} without detailed illustrations. \\
Note: The answer \textbf{must} be in \textbf{\{lang\_name\}}. \\
For example: {\color{tagorange}\texttt{<answer>}} \{example\_answer\} {\color{tagorange}\texttt{</answer>}}. \\

\vspace{5pt}
\hrule
\vspace{5pt}
\textbf{Question:} \{question\}
\end{tcolorbox}
\caption{Multilingual Reasoning and Hierarchical Retrieval Template for \ours. The structure enforces a ``local-first, global-supplement'' strategy to resolve cross-lingual knowledge conflicts.}
\label{table:template_advanced}
\end{table*}

\section{Details on Experimental Results}
\label{app:allResults}
\paragraph{Details on Overall Performance.} Table~\ref{tab:detail_xor} and Table ~\ref{tab:detail_mkqa} show the specific performance of \ours~in all languages on MKQA and XOR-Tydi QA. Results show that methods with retrieval consistently outperform methods with direct inference across nearly all languages. Notably, the performance on Qwen2.5-3B-Instruct generally surpasses that of Qwen3-4B, as the instruct-tuned variant exhibits superior instruction-following capabilities and more robust tag adherence compared to the base models~\cite{jin2025searchr1trainingllmsreason}.
Specifically, Latin-script languages exhibit higher baseline scores and stable performance trajectories, where unseen languages such as \textproc{Es}, \textproc{Pt} and \textproc{De} perform comparably to seen languages like \textproc{Fr} and \textproc{It}, indicating robust intra-family cross-lingual transfer capabilities in current models. Conversely, non-Latin scripts manifest significant performance volatility and pronounced ``performance gaps'' with languages like \textproc{Zh}, \textproc{Ja} and \textproc{Th}, frequently experiencing performance degradation in several baseline models. In extreme low-resource scenarios such as \textproc{Te} and \textproc{Bn}, models face a severe ``low-resource dilemma'' where direct inference often remains nearly optimal, identifying \textproc{Te} as the most challenging language across all evaluated methods. Notably, while \textproc{Ko} and \textproc{Th} exhibits poor performance in unseen languages, it shows substantial improvement potential through retrieval-augmented strategies. Furthermore, RL-based baselines prove highly unstable when generalizing to unseen non-Latin languages, highlighting the inherent difficulty of applying standard RL policies to diverse and complex linguistic scripts.

\begin{figure*}[!t] 
    \centering
    \begin{subfigure}{0.43\linewidth}
        \centering
        \includegraphics[width=\linewidth]{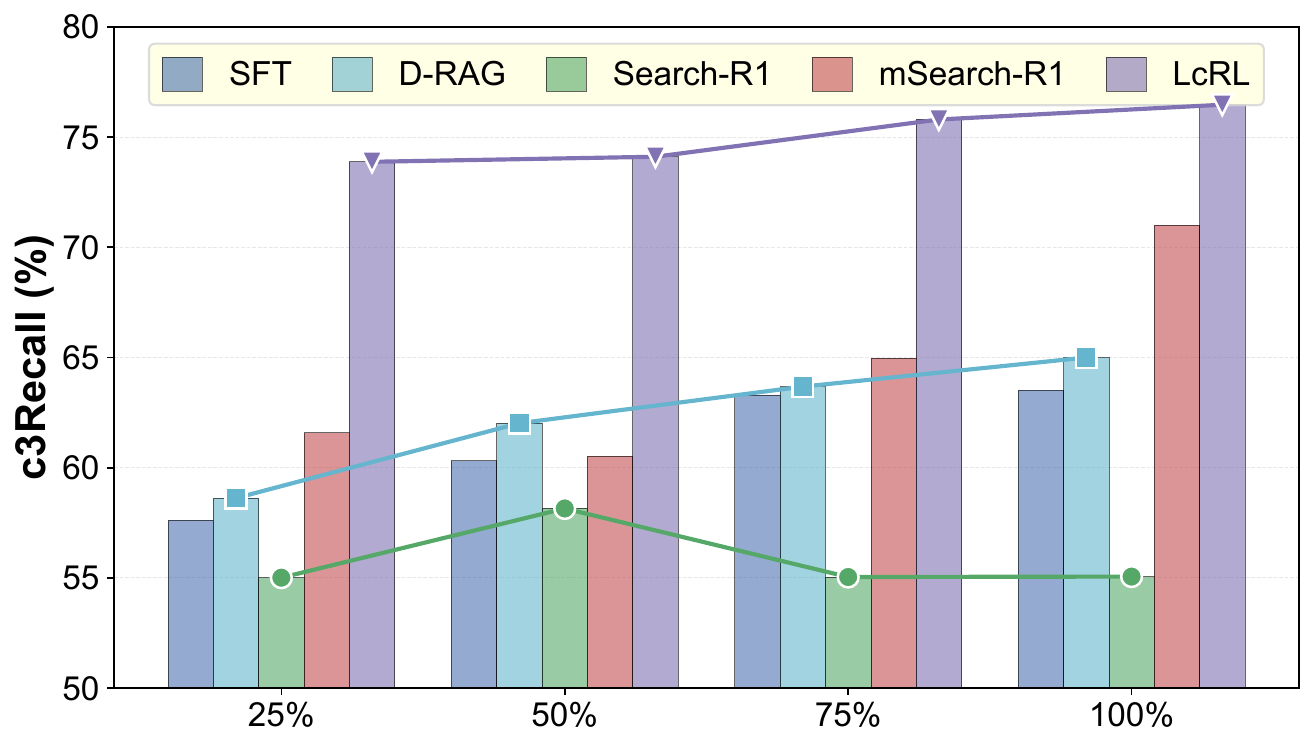} 
        \caption{Performance on Language \textproc{En}.}
        \label{fig:sub21}
    \end{subfigure}
    \begin{subfigure}{0.43\linewidth}
        \centering
        \includegraphics[width=\linewidth]{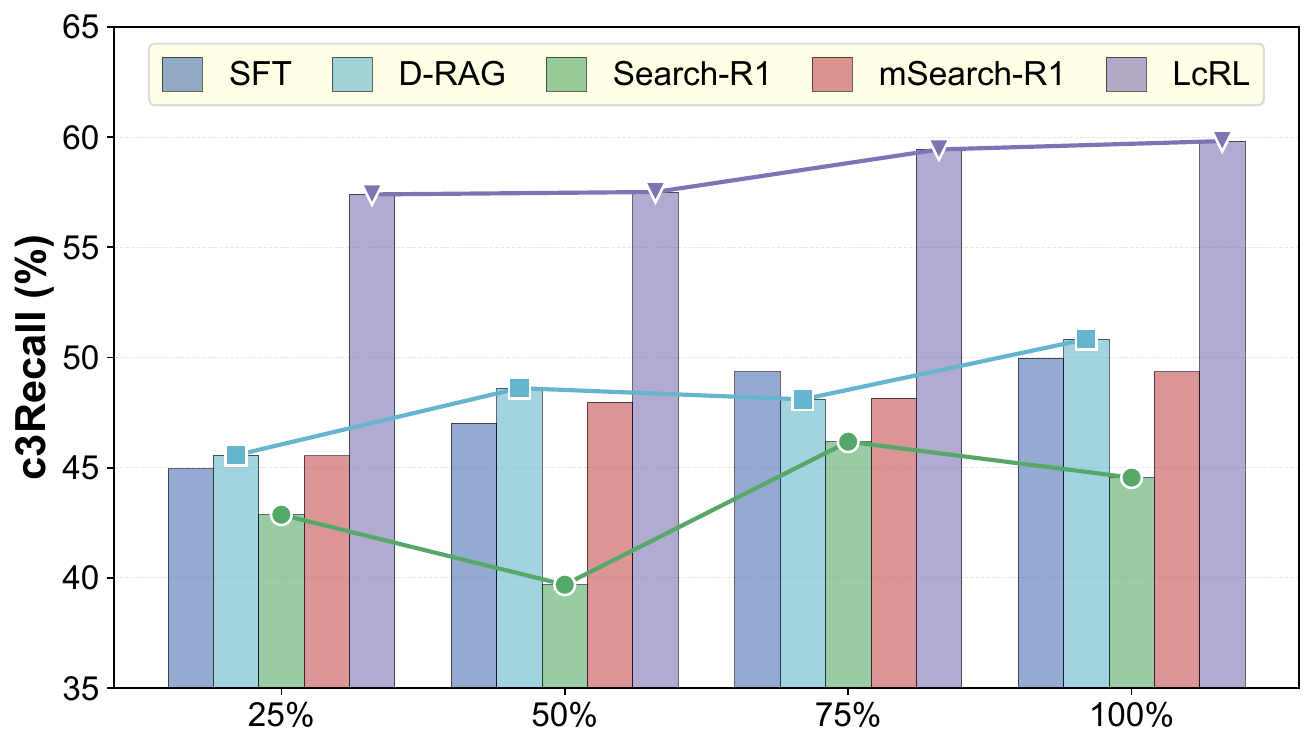} 
        \caption{Performance on Language \textproc{Ru}.}
        \label{fig:sub22}
    \end{subfigure}
        \begin{subfigure}{0.43\linewidth}
        \centering
        \includegraphics[width=\linewidth]{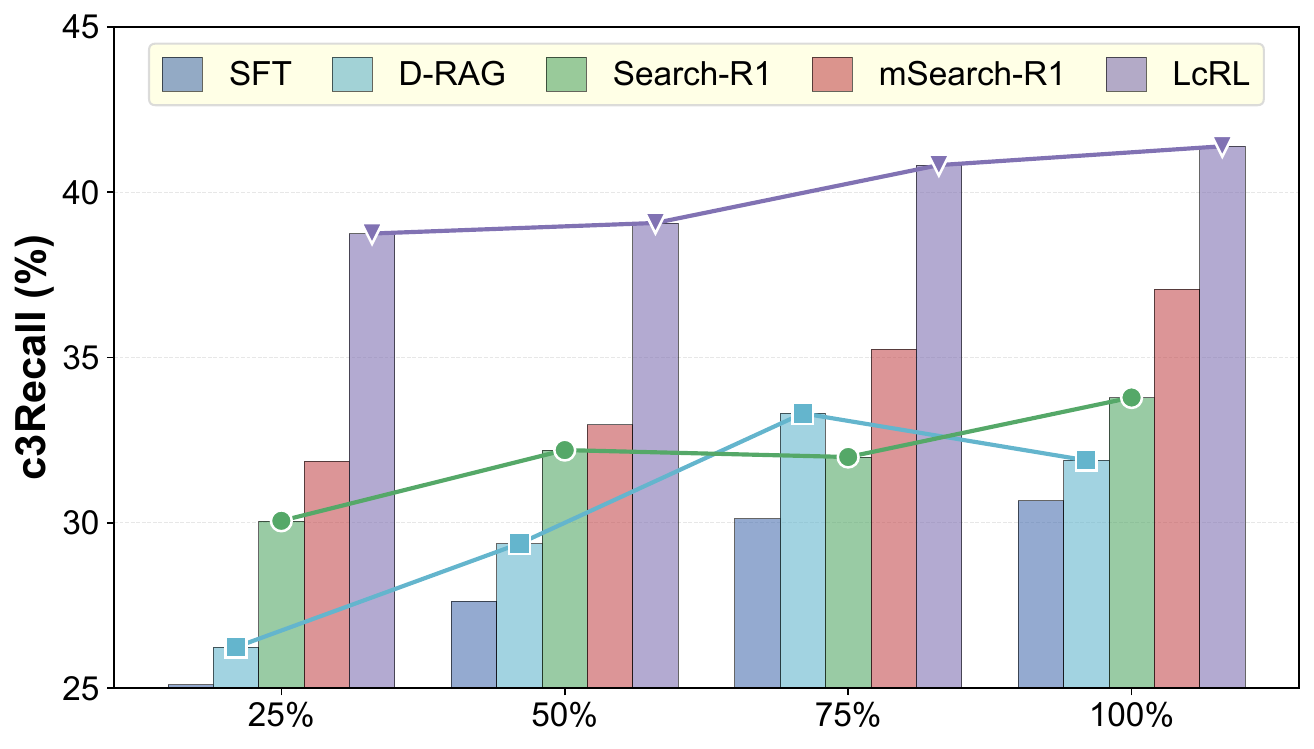} 
        \caption{Performance on Language \textproc{Ar}.}
        \label{fig:sub23}
    \end{subfigure}
    \begin{subfigure}{0.43\linewidth}
        \centering
        \includegraphics[width=\linewidth]{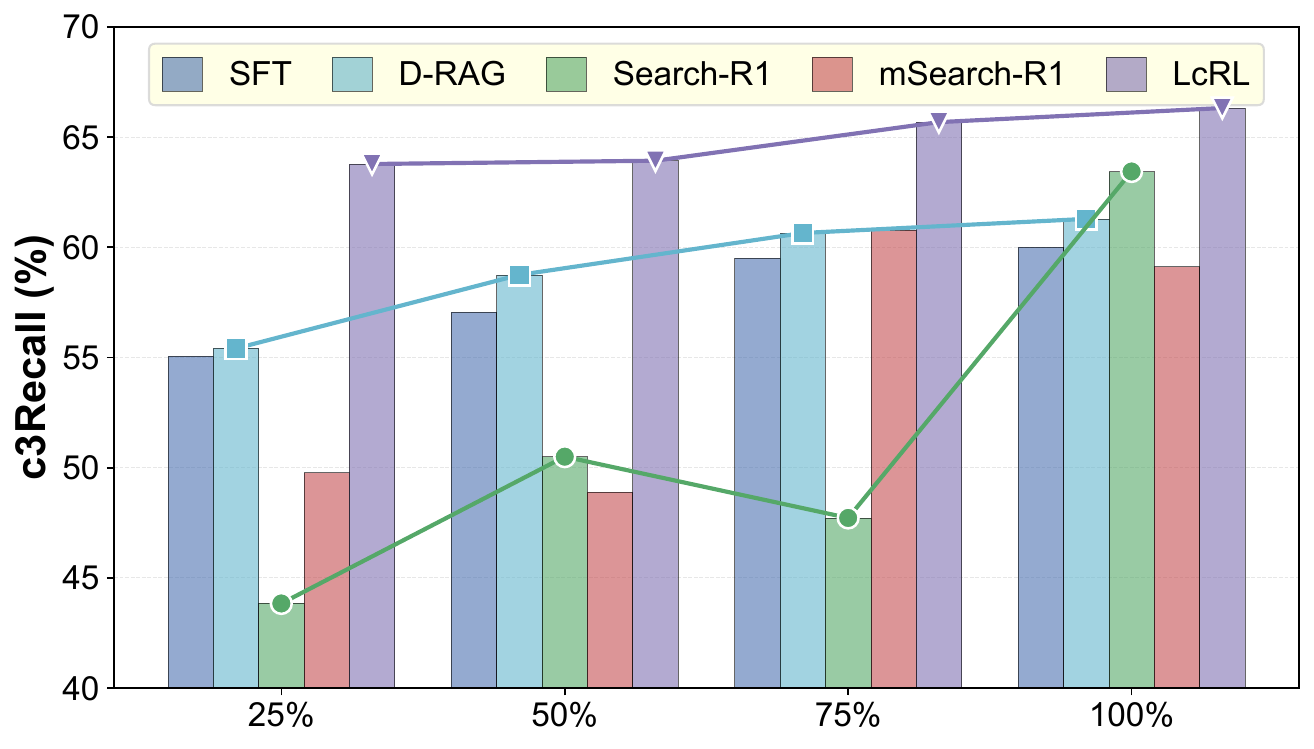} 
        \caption{Performance on Language \textproc{Fr}.}
        \label{fig:sub24}
    \end{subfigure}
        \begin{subfigure}{0.43\linewidth}
        \centering
        \includegraphics[width=\linewidth]{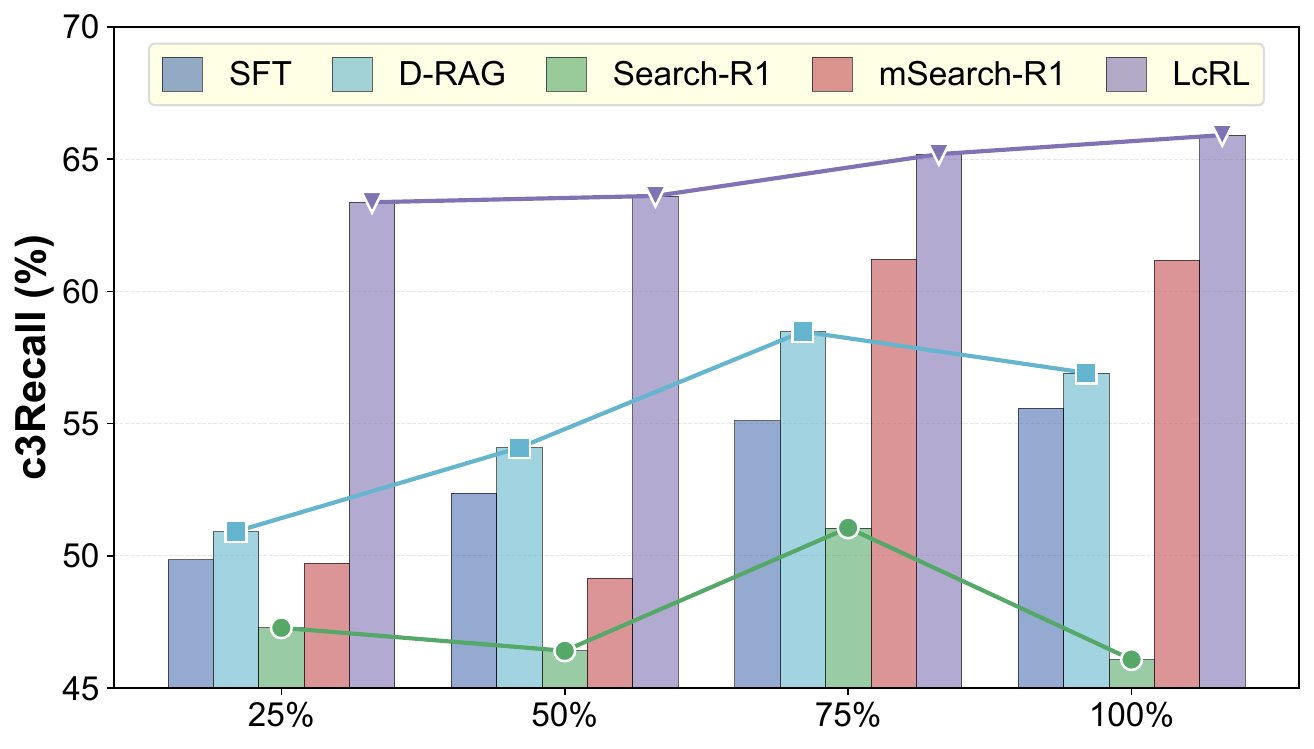} 
        \caption{Performance on Language \textproc{It}.}
        \label{fig:sub25}
    \end{subfigure}
    \begin{subfigure}{0.43\linewidth}
        \centering
        \includegraphics[width=\linewidth]{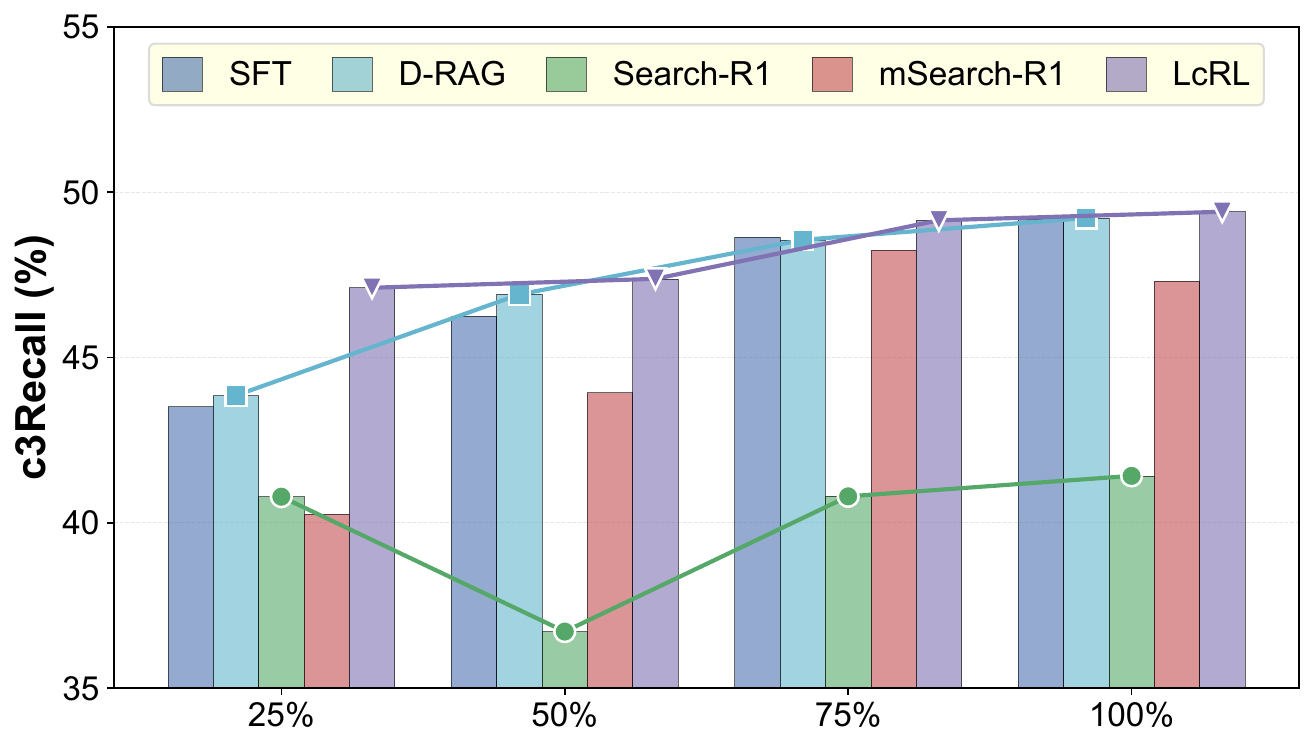} 
        \caption{Performance on Language \textproc{Fi}.}
        \label{fig:sub26}
    \end{subfigure}
        \begin{subfigure}{0.43\linewidth}
        \centering
        \includegraphics[width=\linewidth]{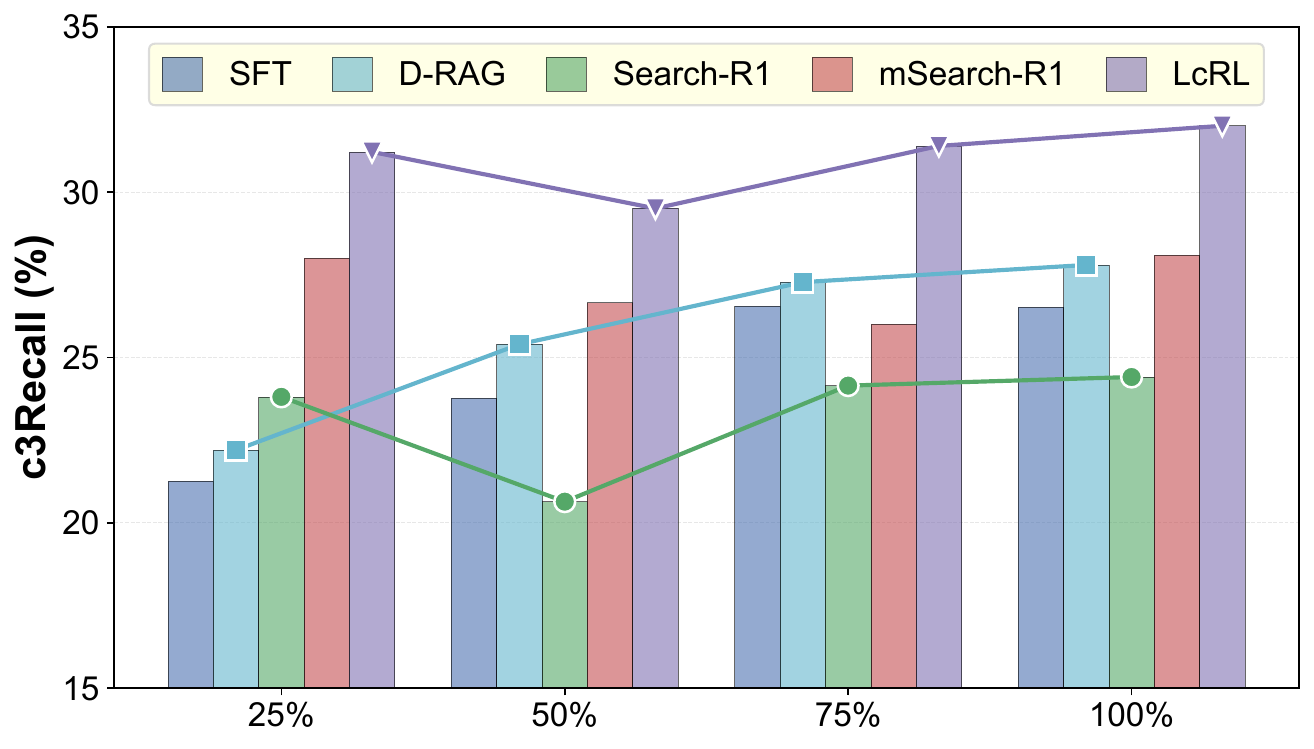} 
        \caption{Performance on Language \textproc{Zh}.}
        \label{fig:sub27}
    \end{subfigure}
    \begin{subfigure}{0.43\linewidth}
        \centering
        \includegraphics[width=\linewidth]{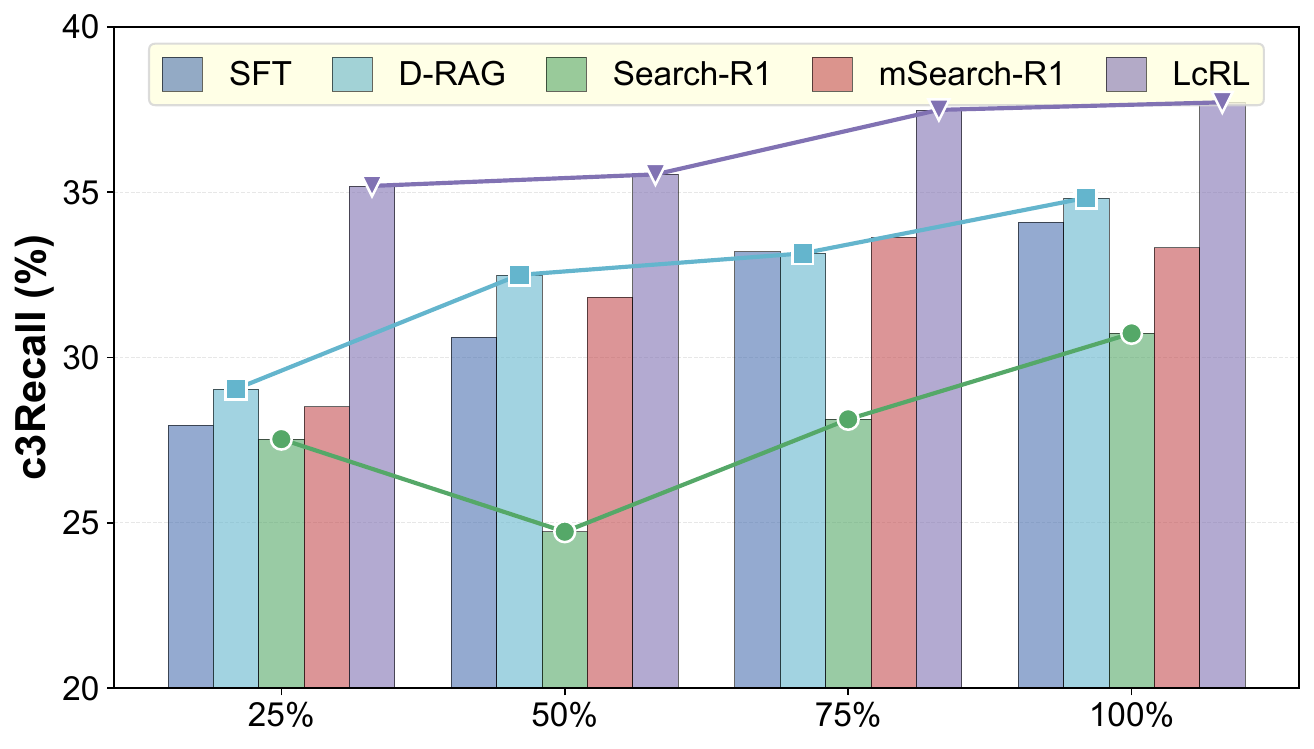} 
        \caption{Performance on Language \textproc{Ja}.}
        \label{fig:sub28}
    \end{subfigure}
    \caption{Results of post-training methods on each language with scaling utilized data.}
    \label{fig:app_scale}
\end{figure*}

\paragraph{Model Sensitivity to Data Scale.} As shown in figure~\ref{fig:app_scale}, by maintaining a fixed number of languages $N$ and scaling the samples per language $M$ from 25\% to 100\% of the total training budget, we observe distinct behavioral patterns:
supervised baselines like SFT and D-RAG exhibit steep linear growth and a heavy dependency on data volume. In contrast, the baseline RL method displays ``unstable'' behavior characterized by performance regressions around the 50\% data scale in languages such as \textproc{Fi}, \textproc{It}, and \textproc{Zh}. Our proposed \ours~demonstrates an ``efficient and robust'' trajectory with only 25\% of the training data, abd it consistently surpasses all baselines trained on the full dataset. 
On high-resource languages such as \textproc{En}, \textproc{Fr}, and \textproc{Ru}, performance across all evaluated methods remains relatively stable. In contrast, on linguistically complex or low-resource languages like \textproc{Ar}, \textproc{Zh}, \textproc{Ja}, and \textproc{Fi}, baseline models exhibit increased volatility, for instance, Search-R1 shows a notable performance decline on Finnish (\textproc{Fi}) as data scales. \ours~maintains high performance starting from the 25\% data volume and exhibits a steady upward trend with the addition of training samples, demonstrating consistent behavior across diverse linguistic settings.

\begin{figure*}[!t] 
    \centering
    \begin{subfigure}{0.43\linewidth}
        \centering
        \includegraphics[width=\linewidth]{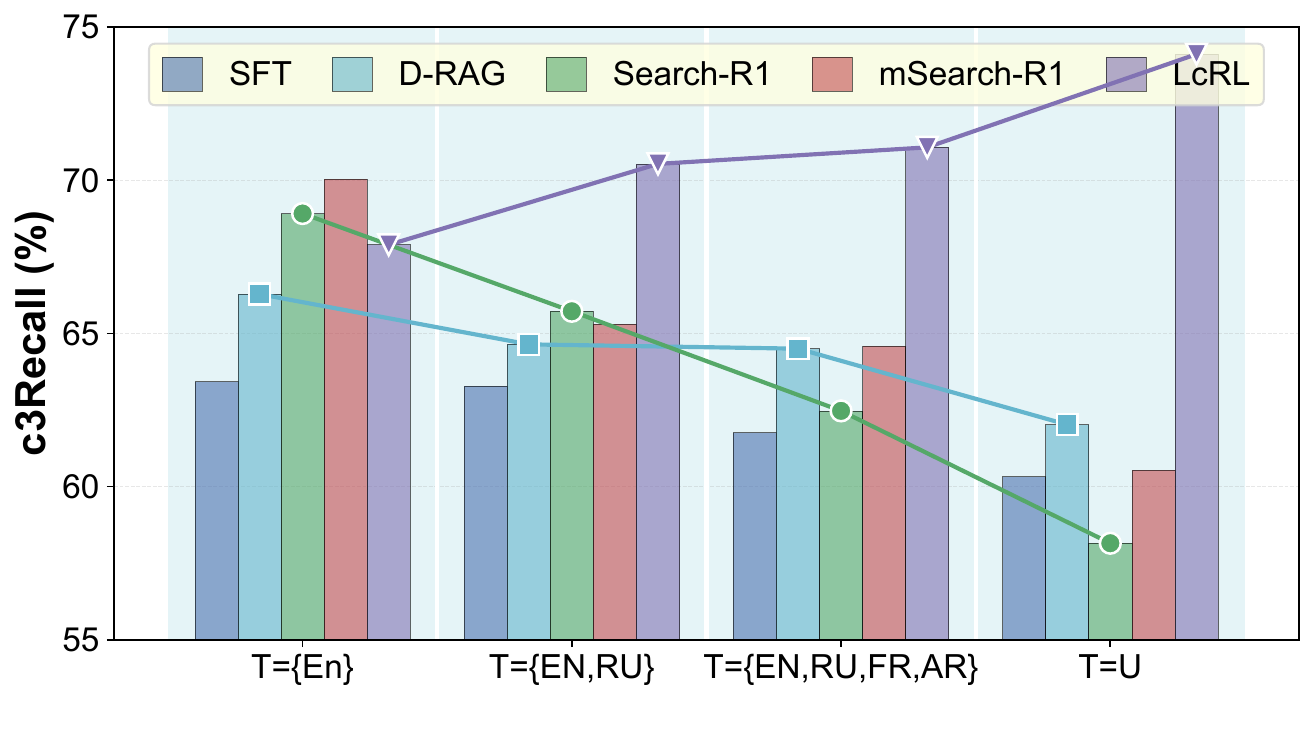} 
        \vspace{-8mm}
        \caption{Performance on Language \textproc{En}.}
        \label{fig:sub11}
    \end{subfigure}
    \begin{subfigure}{0.43\linewidth}
        \centering
        \includegraphics[width=\linewidth]{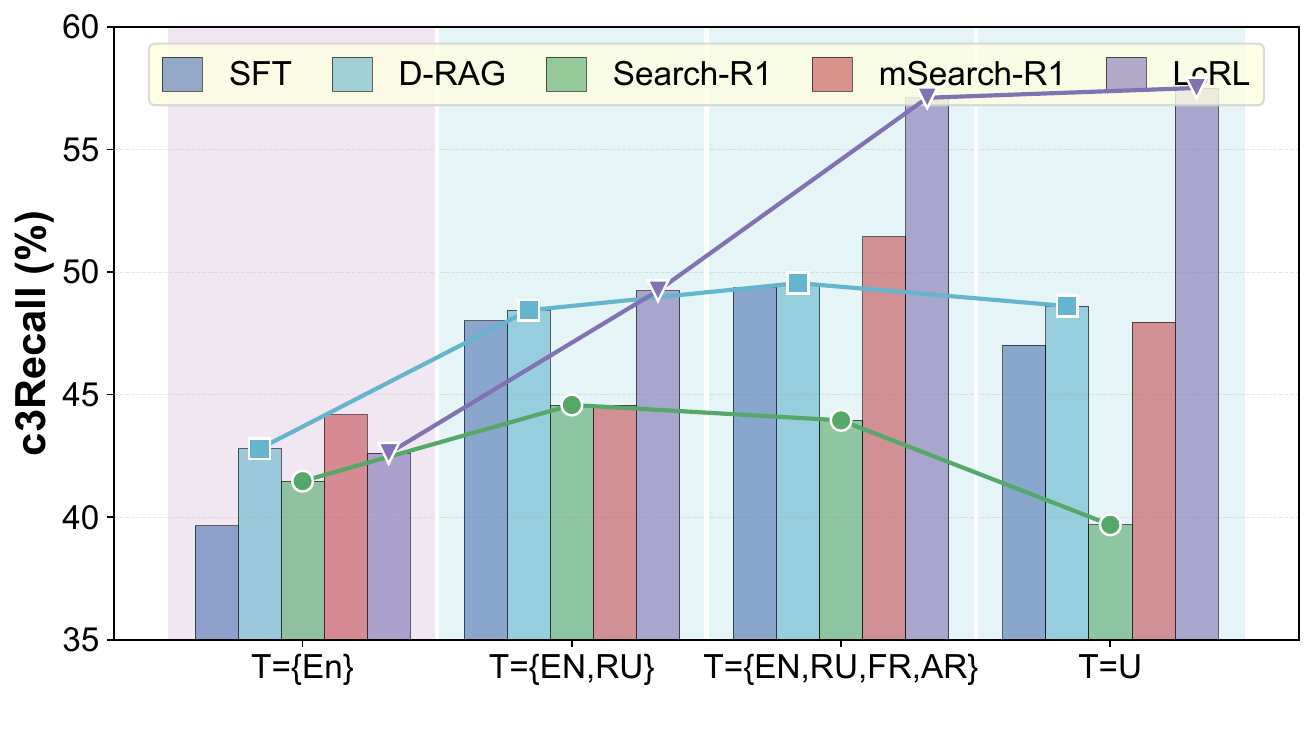} 
        \vspace{-8mm}
        \caption{Performance on Language \textproc{Ru}.}
        \label{fig:sub12}
    \end{subfigure}
        \begin{subfigure}{0.43\linewidth}
        \centering
        \includegraphics[width=\linewidth]{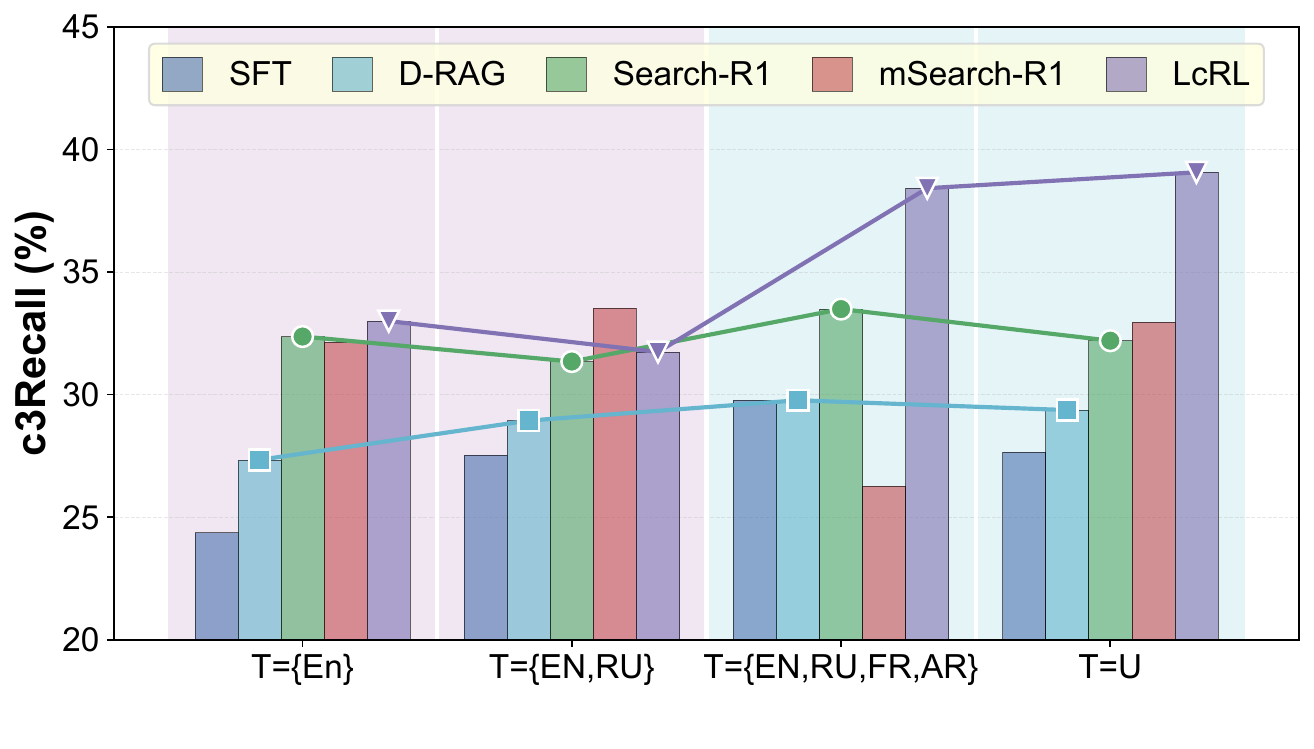} 
        \vspace{-8mm}
        \caption{Performance on Language \textproc{Ar}.}
        \label{fig:sub13}
    \end{subfigure}
    \begin{subfigure}{0.43\linewidth}
        \centering
        \includegraphics[width=\linewidth]{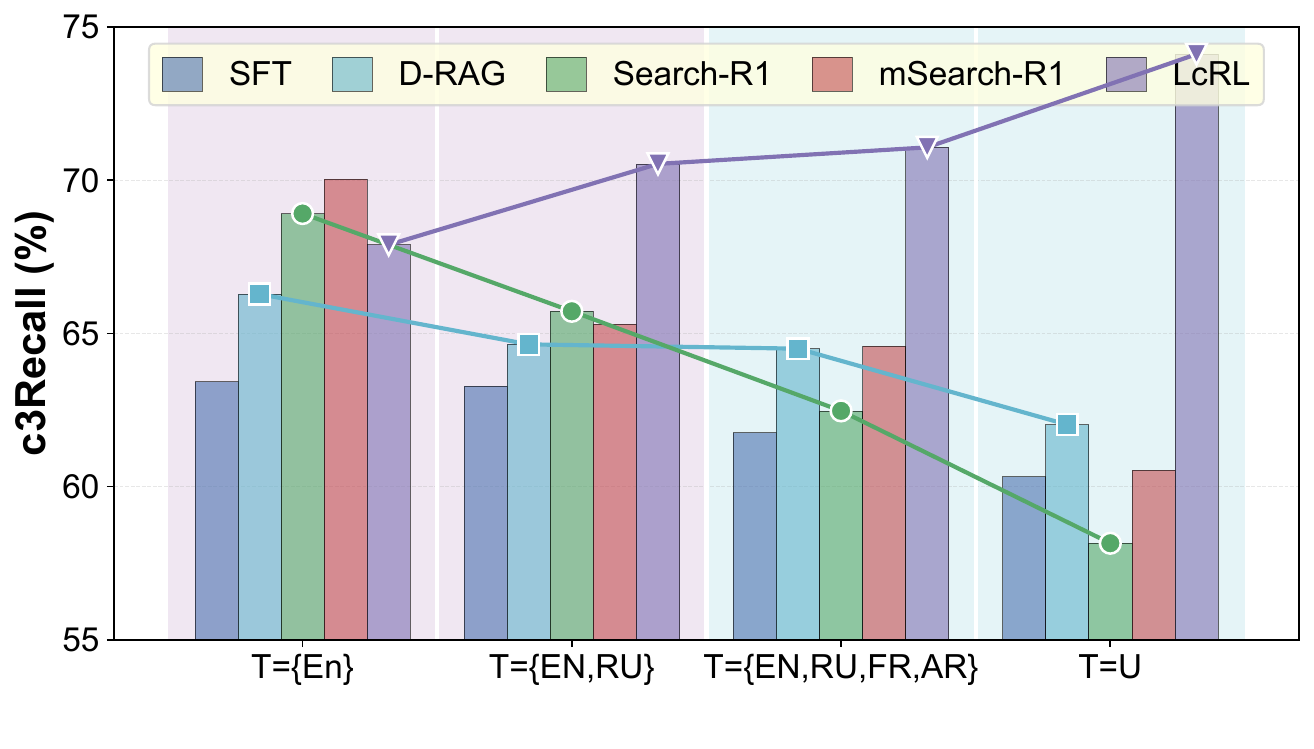} 
        \vspace{-8mm}
        \caption{Performance on Language \textproc{Fr}.}
        \label{fig:sub14}
    \end{subfigure}
        \begin{subfigure}{0.43\linewidth}
        \centering
        \includegraphics[width=\linewidth]{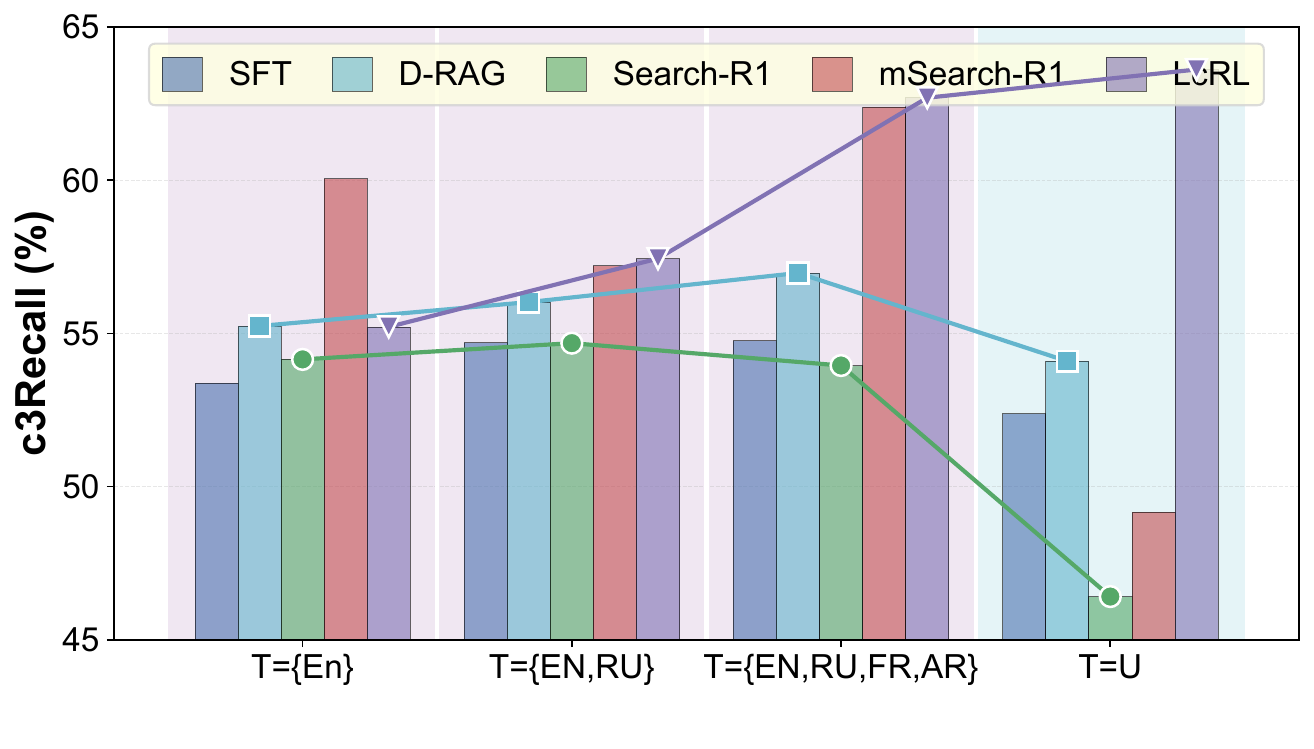} 
        \vspace{-8mm}
        \caption{Performance on Language \textproc{It}.}
        \label{fig:sub15}
    \end{subfigure}
    \begin{subfigure}{0.43\linewidth}
        \centering
        \includegraphics[width=\linewidth]{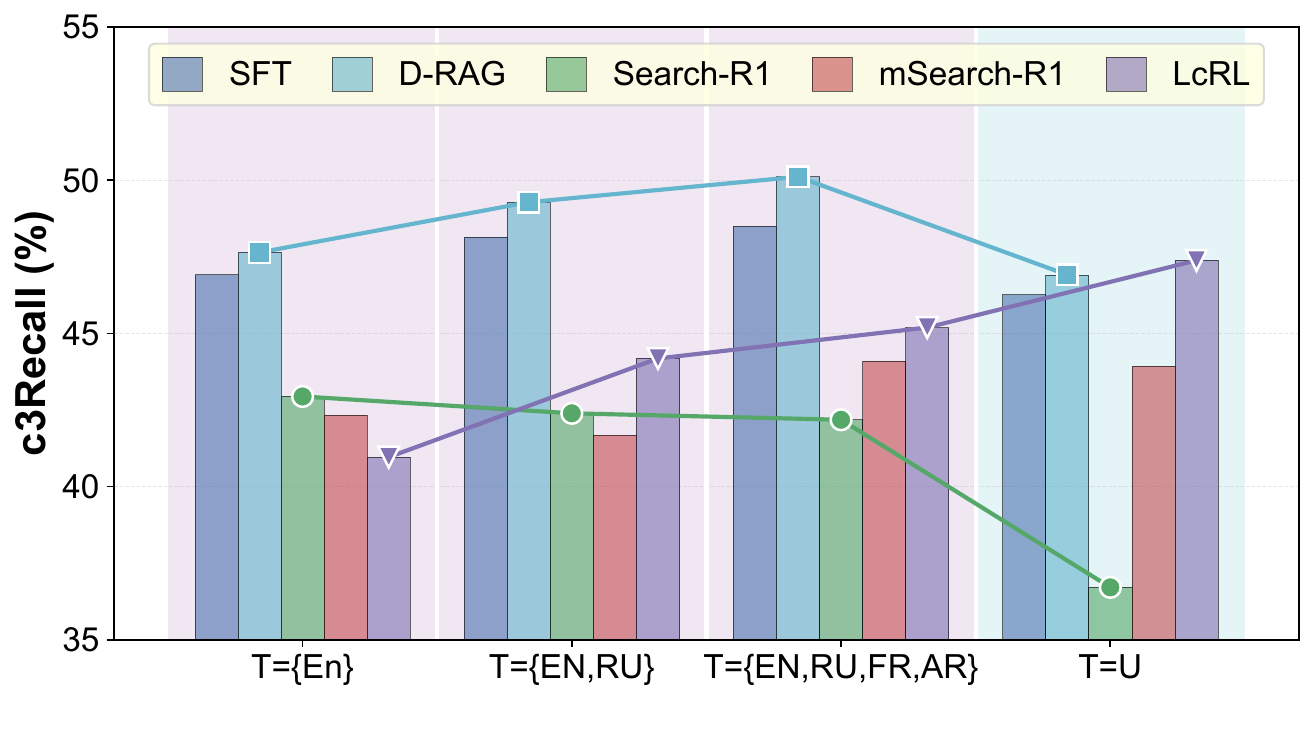} 
        \vspace{-8mm}
        \caption{Performance on Language \textproc{Fi}.}
        \label{fig:sub16}
    \end{subfigure}
        \begin{subfigure}{0.43\linewidth}
        \centering
        \includegraphics[width=\linewidth]{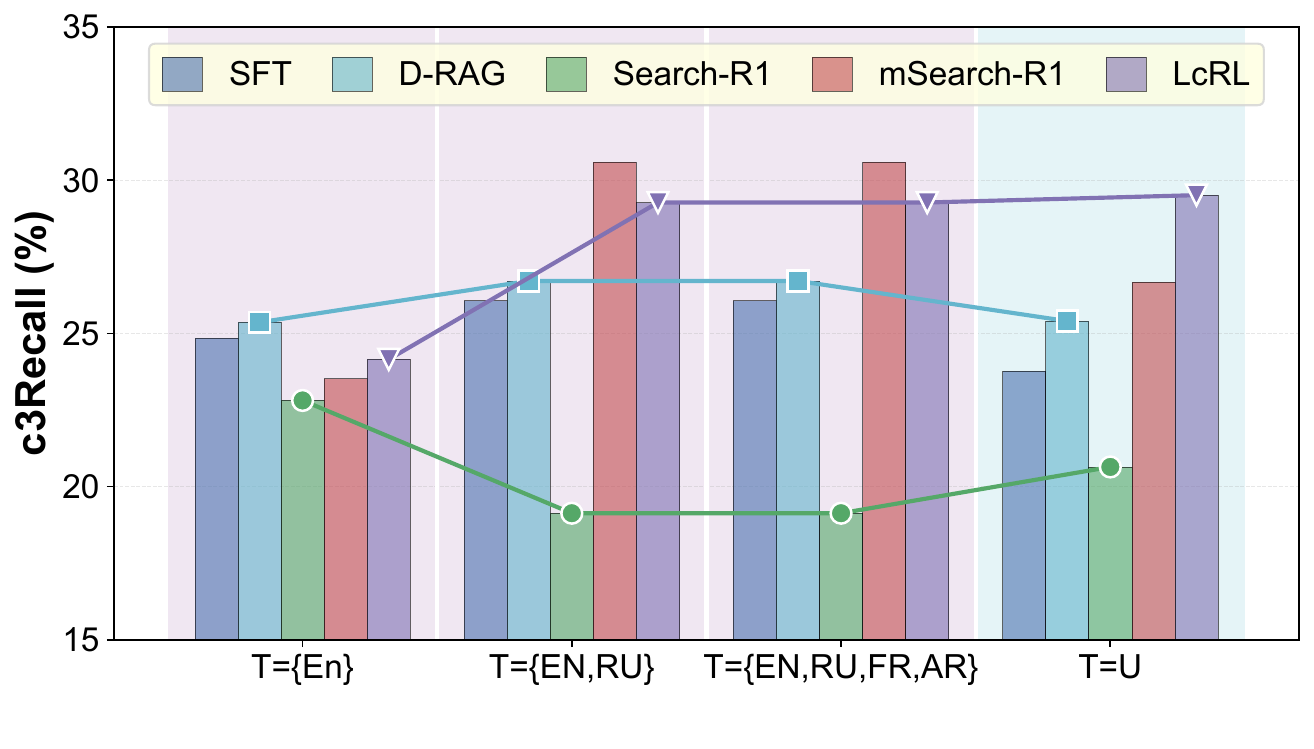} 
        \vspace{-8mm}
        \caption{Performance on Language \textproc{Zh}.}
        \label{fig:sub17}
    \end{subfigure}
    \begin{subfigure}{0.43\linewidth}
        \centering
        \includegraphics[width=\linewidth]{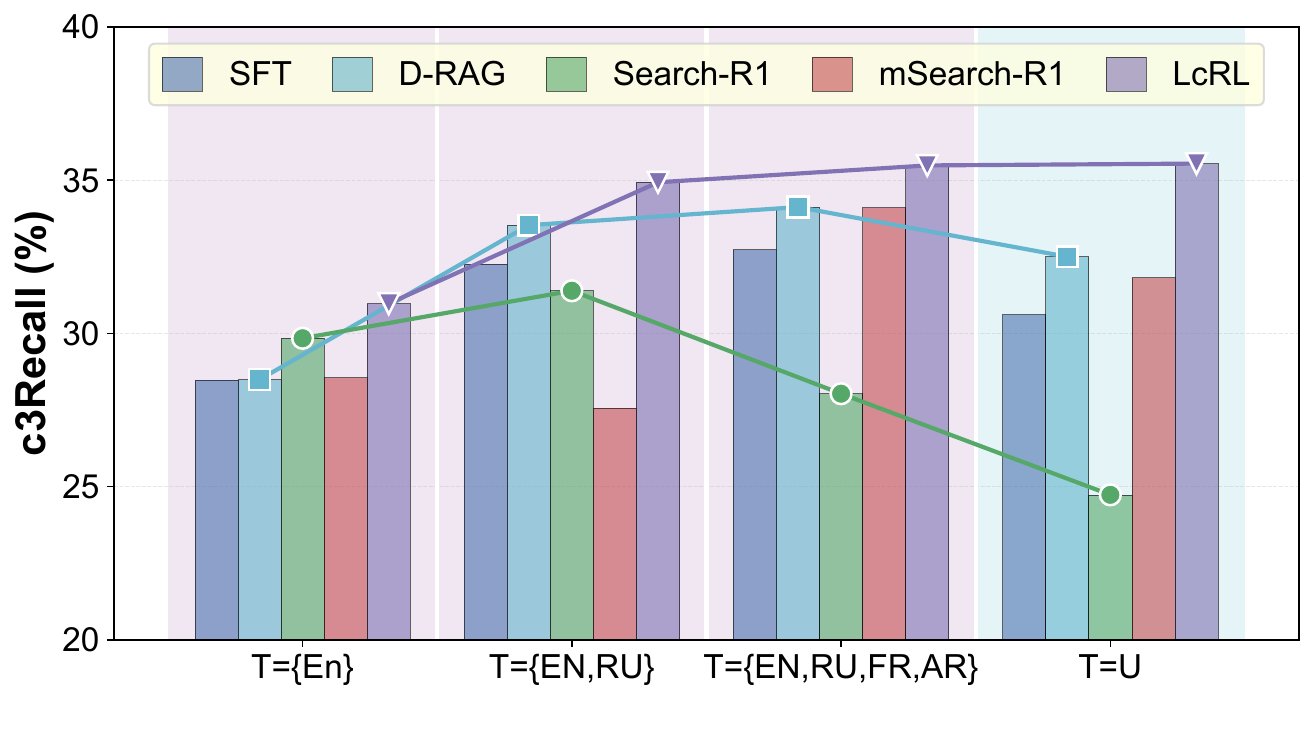} 
        \vspace{-8mm}
        \caption{Performance on Language \textproc{Ja}.}
        \label{fig:sub18}
    \end{subfigure}
    \definecolor{colorGroup1}{RGB}{226, 243, 246}
    \definecolor{colorGroup2}{RGB}{238, 228, 241}
    \caption{Results of post-training methods on each language with different language coverages. Background colors distinguish between \colorbox{colorGroup1}{seen languages} and \colorbox{colorGroup2}{unseen languages}.}
    \label{fig:app_lang}
\end{figure*}

\paragraph{Model Sensitivity to Language Coverage.} As illustrated in Figure~\ref{fig:app_lang}, baseline methods typically exhibit an inverted U-shaped or downward performance trajectory as the number of training languages $N$ increases. This decline is particularly pronounced in high-resource Latin languages like \textproc{En}, \textproc{Fr}, and \textproc{It}, where the inclusion of more languages appears to serve as a hindrance to native performance. Conversely, non-Latin and complex script languages such as \textproc{Zh}, \textproc{Ja}, and \textproc{Ar} maintain relatively stable trends, suggesting a lower susceptibility to cross-lingual interference from additional training data. In unseen language scenarios, baselines often stagnatie or even decline with increasing $N$. Furthermore, RL-based approaches like Search-R1 demonstrate greater sensitivity to data distribution changes with larger performance fluctuations, whereas fine-tuning methods remain comparatively stable. Notably, models consistently achieve effective results even on languages excluded from their specific training sets, underscoring the robust cross-lingual generalization capabilities inherent in these multilingual frameworks.

\section{Training Template}
\label{app:template}
To enable the policy model to autonomously navigate the multilingual knowledge space, we employ a structured template that establishes a clear interaction protocol between the LLM and the search environment. As illustrated in Table~\ref{table:template_advanced}, the template partitions the model’s trajectory into three iterative phases: (1) internal reasoning within <think> tags, (2) tool-calling via <search> tags, and (3) final answer synthesis within <answer> tags. Unlike monolingual reasoning prompts, our template explicitly guides the model to adopt a multi-turn, multi-language retrieval strategy.

\section{Case Study}
\label{appendix:case study}
As shown in Figures \ref{fig:case_study} and \ref{fig:case_study_2}, we conduct a case study on two scenarios involving knowledge bias and knowledge conflict. In the first case (Fig. \ref{fig:case_study}), when asked about the 50th U.S. state, the model recognizes its lack of internal knowledge and initially searches the local Chinese collection. However, it fails to find relevant facts due to knowledge bias, and subsequently triggers a second-round search in global collections. By successfully synthesizing information from English and Japanese documents, the model correctly identifies ``Hawaii'' as the final answer. 
In the second case (Fig. \ref{fig:case_study_2}), when asked about the release time of ``Power of Love'', the model encountered severe knowledge conflict because of multiple distracting dates retrieved from collections in multiple languages. Despite this interference, the model demonstrates robust reasoning by filtering out irrelevant noise and accurately outputting the golden answer (1984). These examples highlight the model's effectiveness in cross-lingual information synthesis and its resilience against contradictory information.

\definecolor{boxborder}{RGB}{0, 0, 0}       
\definecolor{headergray}{RGB}{235, 235, 235} 
\definecolor{hlbluecolor}{RGB}{222, 235, 247} 
\definecolor{hlpurplecolor}{RGB}{230, 225, 245} 
\definecolor{arrowcolor}{RGB}{135, 175, 215}  
\definecolor{orangetext}{RGB}{220, 150, 60}   
\definecolor{hlredcolor}{RGB}{250, 215, 215}
\definecolor{hlgreencolor}{RGB}{215, 240, 215} 



\newcommand{\hlblue}[1]{%
    \bgroup
    \markoverwith{\textcolor{hlbluecolor}{\rule[-0.6ex]{2pt}{2.5ex}}}%
    \ULon{#1}%
}

\newcommand{\hlpurple}[1]{%
    \bgroup
    \markoverwith{\textcolor{hlpurplecolor}{\rule[-0.6ex]{2pt}{2.5ex}}}%
    \ULon{#1}%
}
\newcommand{\hlred}[1]{%
    \bgroup
    \markoverwith{\textcolor{hlredcolor}{\rule[-0.6ex]{2pt}{2.6ex}}}%
    \ULon{#1}%
}

\newcommand{\hlgreen}[1]{%
    \bgroup
    \markoverwith{\textcolor{hlgreencolor}{\rule[-0.6ex]{2pt}{2.6ex}}}%
    \ULon{#1}%
}
\newcommand{\orangeinfo}[1]{%
    \textcolor{orangetext}{\textit{\textbf{#1}}}%
}

\newcommand{\steparrow}{%
    \par\vspace{-2pt}%
    \begin{center}%
        \textcolor{arrowcolor}{\scalebox{1.5}{$\downarrow$}}%
    \end{center}%
    \vspace{-4pt}%
}

\newcommand{\rightarroworange}{%
    \textcolor{arrowcolor}{\Large$\Rightarrow$}%
}


\begin{figure*}[h]
    \centering
    \small
    \begin{tcolorbox}[
        colback=white,          
        colframe=boxborder,     
        boxrule=1.5pt,          
        arc=0pt,                
        left=4pt, right=4pt, top=4pt, bottom=4pt,
        fonttitle=\bfseries
    ]
        
        \begin{tcolorbox}[
            colback=headergray,
            colframe=headergray,
            sharp corners,
            boxsep=0pt, left=2pt, right=2pt, top=2pt, bottom=2pt
        ]
            \textbf{Question:} 美国的第50个州是哪一个? \\
            \textbf{Golden Answer:} 夏威夷州
        \end{tcolorbox}
        
        
        



        
        {\color{tagblue}\texttt{<think>}}\hlblue{美国的第50个州是哪个，我暂时没有这方面的信息，我先进行搜索。}{\color{tagblue}\texttt{</think>}} \\

        \begin{center}
            \vspace{-5pt}
            \textcolor{orangetext}{\textit{\textbf{First search in Chinese collection.}}} \\
            \textcolor{arrowcolor}{\scalebox{1.5}{$\downarrow$}}
            \vspace{-5pt}
        \end{center}
        
        \textbf{Round 1} \\
        {\color{tagpurple}\texttt{<search>}}\hlpurple{美国的第50个州是哪个？}{\color{tagpurple}\texttt{</search>}} \\
        
        {\color{taggreen}\texttt{<information>}} \\
        \texttt{[ZH]} Doc 1(Title: 美国50个州的名称源自不同的语言。) \\
        \texttt{[ZH]} Doc 2(Title: 50号国道总共穿过了12个州，分别是\hlred{加利福尼亚州、内华达州、犹他州}...) \\
        \texttt{[ZH]} Doc 3(Title: 各州同时认可美国宪法，因此其部分的主权被转移给了联邦政府，也就形成由50个共享主权的小国家形成的国家，即美国。) \\
        {\color{taggreen}\texttt{</information>}} \\

        {\color{tagblue}\texttt{<think>}}\hlblue{给出的信息中没有包含“美国的第50个州是哪个？”的相关信息。我需要再次进 \\
        行搜索。}{\color{tagblue}\texttt{</think>}} \\

        \begin{center}
            \vspace{-5pt}
            \textcolor{orangetext}{\textit{\textbf{Next search in global collections.}}} \\
            \textcolor{arrowcolor}{\scalebox{1.5}{$\downarrow$}}
            \vspace{-5pt}
        \end{center}
        
        \textbf{Round 2} \\
        {\color{tagpurple}\texttt{<search>}}\hlpurple{美国的第50个州是哪个？}{\color{tagpurple}\texttt{</search>}} \\
        
        {\color{taggreen}\texttt{<information>}} \\
        \textit{[lang=en]} \\
        \texttt{[EN]} Doc 1(Title: In other fields) Fifty is: There are 50 states in the United States of America. \hlgreen{The TV show Hawaii Five-O and its reimagined version, Hawaii Five-0, are so called because Hawaii is the last (50th) of...}\\ 
        \texttt{[EN]} Doc 2(Title: Route 50S (Ohio–West Virginia) (former)) Alabama State Route 50\\ Arizona State Route 50 (never built)\\ Arkansas Highway 50...\\
        \texttt{[EN]} Doc 3(Title: Route 50) U.S. \\
        
        \textit{[lang=ja]} \\
        \texttt{[JA]} Doc 1(Title: アメリカ合衆国の州は、連邦共和国であるアメリカ合衆国を構成する「国家」で現在50州がある,各州と連邦は国家主権を共有している。) \\
        \texttt{[JA]} Doc 2(Title: 五十州：アメリカ合衆国の州の数。 
    \hlgreen{アメリカ合衆国の50番目の州はハワイ州である。} ローマ数字では 50 は L で表される。 50年祭には、金が記念品として使われたり、形容詞とされることが多い。) \\
        \texttt{[JA]} Doc 3(Title: アメリカ合衆国の50の州 は連邦政府とは主権を共有しながらも独立した準国家統治体である。
        
        \textbf{......} \\
        {\color{taggreen}\texttt{</information>}} \\

        {\color{tagorange}\texttt{<answer>}}美国的第50个州是夏威夷州（Hawaii）。美国一共有50个州，而夏威夷是在1959年正式成为美国联邦的一部分的，因此它是美国最后一个加入联邦的州。{\color{tagorange}\texttt{</answer>}} \\
        


    \end{tcolorbox}
    \caption{Case study for knowledge bias. Green block indicates correct information, and red block indicates relative but incorrect information.}
    \label{fig:case_study}
\end{figure*}

\begin{figure*}[h]
    \centering
    \small
    \begin{tcolorbox}[
        colback=white,          
        colframe=boxborder,     
        boxrule=1.5pt,          
        arc=0pt,                
        left=4pt, right=4pt, top=4pt, bottom=4pt,
        fonttitle=\bfseries
    ]
        
        \begin{tcolorbox}[
            colback=headergray,
            colframe=headergray,
            sharp corners,
            boxsep=0pt, left=2pt, right=2pt, top=2pt, bottom=2pt
        ]
            \textbf{Question:} frankie goes to hollywood power of love 是哪一年的? \\
            \textbf{Golden Answer:} 1984
        \end{tcolorbox}

        {\color{tagblue}\texttt{<think>}}\hlblue{frankie goes to hollywood power of love 是哪一年的?\\我需要通过搜索引擎来查找相关信息。}{\color{tagblue}\texttt{</think>}} \\

        \begin{center}
            \vspace{-5pt}
            \textcolor{orangetext}{\textit{\textbf{First search in Chinese collection.}}} \\
            \textcolor{arrowcolor}{\scalebox{1.5}{$\downarrow$}}
            \vspace{-5pt}
        \end{center}
        
        \textbf{Round 1} \\
        {\color{tagpurple}\texttt{<search>}}\hlpurple{frankie goes to hollywood power of love 发行时间}{\color{tagpurple}\texttt{</search>}} \\
        
        {\color{taggreen}\texttt{<information>}} \\
        \texttt{[ZH]} Doc 1(Title: 影片改期到2021年9月3日在美国院线发行，海外发行于2021年8月25日。) \\
        \texttt{[ZH]} Doc 2(Title: 据报道，发行商焦点影业将该片计划于2018年8月17日在美国上映。 之后，美国上映日被改至2019年3月29日；不久后，上映日又提前为2019年3月15日。) \\
        \texttt{[ZH]} Doc 3(Title: 影片于2022年7月15日由派拉蒙影業在北美地区发行，当地原定档期为2017年4月14日，后来推迟到2017年8月4日。) \\
        {\color{taggreen}\texttt{</information>}} \\

        {\color{tagblue}\texttt{<think>}}\hlblue{我需要搜索更多信息来给出答案。}{\color{tagblue}\texttt{</think>}} \\

        \begin{center}
            \vspace{-5pt}
            \textcolor{orangetext}{\textit{\textbf{Next search in global collections.}}} \\
            \textcolor{arrowcolor}{\scalebox{1.5}{$\downarrow$}}
            \vspace{-5pt}
        \end{center}
        
        \textbf{Round 2} \\
        {\color{tagpurple}\texttt{<search>}}\hlpurple{when was the song power of love by frankie released}{\color{tagpurple}\texttt{</search>}} \\
        
        {\color{taggreen}\texttt{<information>}} \\
        \textit{[lang=en]} \\
        \texttt{[EN]} Doc 1(Title: It also topped the charts in several other European countries...)\\ 
        \texttt{[EN]} Doc 2(Title: ``The Power of Love'' is a song originally recorded and released by British band Frankie Goes to Hollywood. It was written by...) \\
        \texttt{[EN]} Doc 3(Title: ``Bat Out of Hell'' originally peaked outside the top ten at number 15 upon \hlred{its initial release in 1979.} ``The Power of Love'' originally peaked at number-one upon \hlgreen{its initial release in 1984.})\\
        
        \textit{[lang=it]} \\
        \texttt{[IT]} Doc 1(Title: The Power of Love è il terzo singolo estratto dall'album Welcome to the Pleasuredome dei Frankie Goes to Hollywood, \hlgreen{pubblicato nel 1984.}) \\
        \texttt{[IT]} Doc 2(Title: Power of Love è il settimo album in studio del cantautore statunitense Luther Vandross, \hlred{pubblicato nel 1991. Tracce} ) \\
        \texttt{[IT]} Doc 3(Title: \hlred{Nel giugno 1985,} The Power of Love venne pubblicato come singolo nel Regno Unito, dove scalò la classifica britannica, rimanendo in prima posizione \hlred{per cinque settimane nell'ottobre 1985} e diventando...)
        
        \textbf{......} \\
        {\color{taggreen}\texttt{</information>}} \\

        {\color{tagblue}\texttt{<think>}} 根据搜索结果，Frankie Goes to Hollywood 的歌曲《Power of Love》最初是在1984年发行的。这首歌在美国取得了巨大的成功，并且在英国排行榜上也表现优异，最终在1985年达到了榜首位置。 虽然有报道称该歌曲计划在2018年和2019年在美国上映，但这些信息似乎与歌曲的发行时间无关。因此，《Power of Love》最准确的发行年份是1984年。{\color{tagblue}\texttt{</think>}}\\
        {\color{tagorange}\texttt{<answer>}}《Power of Love》最准确的发行年份是1984年。{\color{tagorange}\texttt{</answer>}} \\

    \end{tcolorbox}
    \caption{Case study for knowledge conflict. Green block indicates correct information. Red block indicates relative but incorrect information. Purple block indicates no relative information.}
    \label{fig:case_study_2}
\end{figure*}

\end{CJK*}
\end{document}